\documentclass{article}
\usepackage{iclr2027_conference,times}

\usepackage{amsmath,amsfonts,bm}

\def\eqref#1{equation~\ref{#1}}

\def\1{\bm{1}}

\DeclareMathAlphabet{\mathsfit}{\encodingdefault}{\sfdefault}{m}{sl}
\SetMathAlphabet{\mathsfit}{bold}{\encodingdefault}{\sfdefault}{bx}{n}

\usepackage{amsmath,amssymb,amsfonts}
\newtheorem{proposition}{Proposition}
\newtheorem{corollary}{Corollary}
\usepackage{graphicx}
\usepackage{booktabs}
\usepackage{multirow}
\usepackage{wrapfig}
\usepackage{url}
\usepackage{xcolor}
\usepackage{colortbl}
\definecolor{ReseeBlue}{HTML}{3B82E0}
\definecolor{ReseeBand}{HTML}{E6ECF7}
\definecolor{ReseeRed}{HTML}{E0533C}
\usepackage[most]{tcolorbox}
\definecolor{ReseeOrange}{HTML}{EE8A2C}
\usepackage{enumitem}
\newtcolorbox{promptbox}[2][]{enhanced, breakable, arc=1mm, outer arc=1mm, boxrule=0.5pt,
  colframe=black!30, colback=white, colbacktitle=black!5, coltitle=black!85,
  fonttitle=\small\bfseries, title={#2}, toptitle=2.5pt, bottomtitle=2.5pt,
  left=6pt, right=6pt, top=4pt, bottom=4pt, before skip=9pt, after skip=9pt,
  fontupper=\footnotesize, before upper={\setlength{\parskip}{3pt}}, #1}
\newtcolorbox{minicard}[2]{enhanced, arc=1mm, boxrule=0.4pt, colframe=black!25, colback=white,
  title={\textcolor{#1}{#2}}, fonttitle=\scriptsize\bfseries, coltitle=black!85, colbacktitle=black!4,
  toptitle=1.5pt, bottomtitle=1.5pt, left=5pt, right=5pt, top=3pt, bottom=3pt, fontupper=\footnotesize}
\newcommand{\ptag}[3]{{\ttfamily\small\color{#2}#3}}
\newcommand{\tlook}[1]{\ptag{ReseeSlate}{ReseeSlate!80!black}{#1}}
\newcommand{\trecover}[1]{\ptag{ReseeBlue}{ReseeBlue!70!black}{#1}}
\newcommand{\tanswer}[1]{\ptag{ReseeGreen}{ReseeGreen!70!black}{#1}}
\newcommand{\tthink}[1]{\ptag{ReseePurple}{ReseePurple!75!black}{#1}}
\definecolor{ReseeGreen}{HTML}{23A047}
\definecolor{ReseePurple}{HTML}{7C5CBF}
\definecolor{ReseeSlate}{HTML}{5B6472}
\usepackage{hyperref}

\newcommand{\care}{\textsc{Care}}
\newcommand{\resee}{\textsc{ReSee}}
\newcommand{\tagname}[1]{\texttt{<#1>}}
\usepackage{algorithm}
\usepackage{algpseudocode}
\algrenewcommand\algorithmicrequire{\textbf{\color{ReseeBlue!70!black}Input:}}
\algrenewcommand\algorithmicensure{\textbf{\color{ReseeBlue!70!black}Output:}}
\algrenewcommand{\algorithmiccomment}[1]{\hfill{\color{ReseeBlue!80!black}\footnotesize$\triangleright$ #1}}
\newcommand{\algphase}[1]{\Statex {\color{ReseeOrange!85!black}\textit{// #1}}}
\newtcolorbox{trajbox}[3][]{enhanced, arc=1mm, outer arc=1mm, boxrule=0.5pt,
  colframe=black!30, colback=white, colbacktitle=#2!6, coltitle=black!85,
  fonttitle=\small\bfseries, title={#3}, toptitle=2.5pt, bottomtitle=2.5pt,
  left=6pt, right=6pt, top=4pt, bottom=4pt, before skip=8pt, after skip=6pt,
  fontupper=\footnotesize, before upper={\setlength{\parskip}{2.5pt}}, #1}
\newtcolorbox{trrow}[2][white]{blanker, left=0pt, right=0pt, top=2.8pt, bottom=2.8pt,
  before skip=0pt, after skip=0pt, borderline north={0.4pt}{0pt}{black!12},
  fontupper=\footnotesize}
\newcommand{\trstep}[2]{\textcolor{#1}{\rule[0.15ex]{0.42em}{0.42em}}\hspace{0.4em}%
  {\color{#1}\scshape\bfseries #2}\hspace{0.5em}}
\definecolor{TrAsk}{HTML}{55606E}
\definecolor{TrPlan}{HTML}{7A6BA6}
\definecolor{TrLook}{HTML}{4A7FB5}
\definecolor{TrLookB}{HTML}{6E9BC4}
\definecolor{TrRex}{HTML}{C08A3E}
\definecolor{TrRexB}{HTML}{A8712C}
\definecolor{TrOk}{HTML}{4A8B5C}
\definecolor{TrBad}{HTML}{B5615C}
\definecolor{ReseeSuccess}{HTML}{2F7BE0}
\definecolor{ReseeFail}{HTML}{E0413C}
\title{Toward Counterfactual Learning for Selective Re-Examination in Agentic Visual Search}

\author{Ao Chen$^{1}$, Shengchao Chen$^{2*\dagger}$, Yan Luximon$^{1*}$ \\
$^{1}$The Hong Kong Polytechnic University \quad $^{2}$University of Technology Sydney \\
\texttt{aosir.chen@connect.polyu.hk}, \texttt{shengchao.chen.uts@gmail.com}, \\
\texttt{yan.luximon@polyu.edu.hk}
}

\iclrfinalcopy

\begin{document}
\maketitle
\lhead{}
\renewcommand{\headrulewidth}{0pt}
{\renewcommand{\thefootnote}{\fnsymbol{footnote}}\footnotetext[1]{Corresponding authors. $^{\dagger}$Project leader.}}

\begin{abstract}
Visual search over high-resolution images requires a vision--language model (VLM) not only to determine \emph{where} to inspect, but also \emph{whether} another observation is warranted before committing to an answer. Existing prompting, demonstrations, and outcome-based reinforcement learning can encourage re-examination, yet do not directly supervise whether it is beneficial at a particular search state. We introduce \resee{}, a benchmark of re-examination decisions comprising $3{,}227$ visual-search trajectories and $8{,}827$ explicit commit-or-re-examine decisions, with counterfactual adjudication that distinguishes states where additional inspection is warranted from those where commitment is preferable. Building on this formulation, we propose \care{} (\textbf{C}ounterfactual \textbf{A}dvantage for \textbf{R}e-\textbf{E}xamination), a decision-level reinforcement learning method that compares paired continuations under commit and re-examine verdicts from the same search state. Their cost-aware return difference estimates the counterfactual value of an additional observation and is assigned only to the corresponding verdict tokens, providing localized credit for whether search should continue. Experiments expose limited re-examination selectivity in existing VLMs and show that \care{} significantly improves accuracy and decision selectivity over GRPO across two backbones, with higher accuracy and fewer visual requests on six unseen OOD benchmarks.
\end{abstract}

\section{Introduction}
\label{sec:intro}

Visual search is a core capability for vision--language models (VLMs) operating on real images, where the evidence required to answer a question may occupy only a small fraction of a high-resolution frame~\citep{zhang2025mme}. Recent work therefore formulates visual understanding as a multi-step interaction in which the model selects a region, observes a crop, updates its belief, and decides where to inspect next~\citep{zheng2026deepeyes,lai2026mini}. This interaction introduces a trade-off. Committing too early can lock in errors before sufficient evidence has been gathered, whereas unnecessary inspection consumes a finite interaction budget. Agentic VLMs now support crop, zoom, and backtrack actions within a single episode~\citep{wang2026traceable,sarch2026grounded}, making this trade-off central to long-horizon visual search. Beyond learning \emph{where} to look, a visual search policy must therefore also determine \emph{whether an additional observation is warranted}.

\begin{figure}[tbh]
\centering
\includegraphics[width=.98\textwidth]{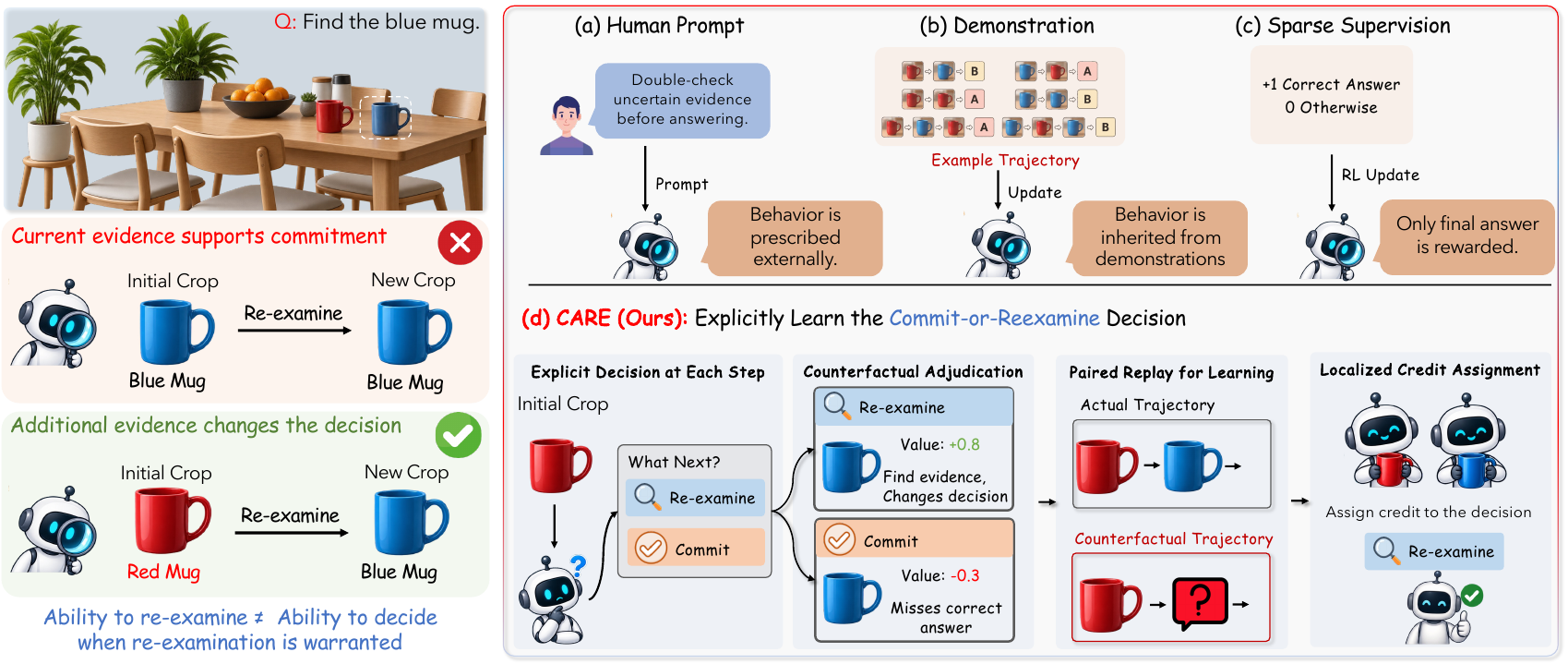}
\vspace{-8pt}
\caption{Re-examination as an explicit decision. The left panel contrasts warranted and unwarranted re-examination. The right panel shows how \care{} evaluates the commit or re-examine verdict through paired continuations and credits that verdict directly.}
\label{fig:intro}
\vspace{-14pt}
\end{figure}

Existing visual search agents are largely optimized through trajectory-level outcomes~\citep{zheng2026deepeyes,su2026pixel,lai2026mini}. Reinforcement learning with answer-based rewards can encourage a policy to locate informative regions, interpret their content, and ultimately produce a correct, grounded answer. Such supervision, however, does not identify the marginal value of an individual observation. A trajectory in which re-examination corrects an erroneous answer and one in which it merely confirms an already-correct answer may receive the same final reward. Prompted verification~\citep{madaan2023self,huang2024large}, explicit rewards for additional checking~\citep{su2026pixel}, and demonstrations that include re-examination can all encourage the behavior itself, but none directly supervises whether re-examination is warranted at a particular search state (Figure~\ref{fig:intro}a--c). Consequently, the ability to re-examine does not imply the ability to decide \emph{whether} re-examination is warranted at the current search state (Figure~\ref{fig:intro}, left).

This limitation points to a more precise learning target: the state-dependent value of re-examination. At each search state, the desired signal should quantify the benefit of acquiring an additional observation relative to committing with the evidence already available. Trajectory-level answer rewards cannot provide this supervision~\citep{kazemnejad2024vineppo,feng2026group}, because re-examinations with different counterfactual utility may nevertheless receive the same final reward. Nor can the policy's own confidence or uncertainty directly supply this value, as such signals describe its current belief rather than the counterfactual effect of observing additional evidence. Learning this decision therefore requires both an explicit representation of the commit-or-re-examine choice and supervision that attributes the consequence of an additional observation directly to that choice.

To address this gap, we introduce \resee{}, a benchmark of explicit commit-or-reexamine decisions, and \care{}, a learning method for assigning decision-level credit to re-examination (Figure~\ref{fig:intro}d). \resee{} contains trajectories in which a conservatively searching VLM records a commit-or-reexamine verdict at each decision state, exposing both decisions to continue searching and decisions to commit. Rather than treating the model's verdict as supervision, we counterfactually adjudicate each state by comparing the consequences of re-examination and immediate commitment, yielding cases where additional inspection is beneficial and where commitment is preferable. Building on these decisions, \care{} constructs paired continuations under the two verdicts from the same search state and uses their cost-adjusted return difference to estimate the value of re-examination. This advantage is assigned only to the verdict tokens that determine whether search continues, rather than to the subsequent trajectory, providing localized credit for  commit-or-reexamine decision. Contributions:
\begin{itemize}
\item We formulate re-examination in long-horizon visual search as a decision-level credit-assignment problem, with the learning target defined as the counterfactual value of an additional observation relative to immediate commitment.

\item We introduce \resee{}, a benchmark of re-examination decisions comprising $3{,}227$ visual-search trajectories and $8{,}827$ explicit commit-or-re-examine decisions, including $1{,}862$ counterfactually adjudicated cases. Its positive and negative cases separate policy self-assessment from the measured utility of re-examination.

\item We propose \care{}, a decision-level RL method that constructs paired continuations under opposite verdicts and assigns their cost-aware return difference only to the corresponding verdict tokens, providing localized credit for the commit-or-re-examine decision.

\item Experiments on \resee{} expose limited decision selectivity in existing VLMs. Across two backbones, \care{} improves macro accuracy and balanced decision accuracy over GRPO with fewer visual requests. Further analyses show accuracy gains on six unseen OOD benchmarks and across a range of model capacities.
\end{itemize}

\section{Related Work}
\label{sec:related}

\paragraph{Long-Horizon Visual Search.}
Long-horizon visual search extends visual question answering beyond single-view inference by acquiring evidence through sequential interaction. Early approaches guide frozen VLMs through targeted or tree-structured exploration~\citep{wu2024vstar,shen2025zoomeye}, while grounded chain-of-thought supervision associates intermediate reasoning with relevant image regions~\citep{shao2024visual}. More recent agentic VLMs learn the search process itself, using multi-turn crop or zoom actions~\citep{zheng2026deepeyes,su2026pixel,lai2026mini}, grounding reasoning in image coordinates or traceable evidence regions~\citep{sarch2026grounded,wang2026traceable}, or delegating localization to dedicated search agents~\citep{li2025insight}. Many such systems optimize trajectory-level correctness with verifiable outcome rewards and group-relative reinforcement learning objectives. Recent analysis further suggests that crop-and-zoom training improves intrinsic reasoning more readily than it enables additional observations to correct intrinsic failures~\citep{ma2026does}. Interaction efficiency is typically handled separately through fixed budgets or mechanisms that regulate computation, including length penalties and adaptive allocation based on the expected return of further reasoning~\citep{chen2024not,lyu2025hierarchical,zhou2026nice}. These approaches regulate search effort at the trajectory or budget level, but do not directly estimate the state-dependent value of an additional visual observation.

\paragraph{Self-Correction and Credit Assignment.}
Self-correction encourages models to verify, critique, and revise their outputs through prompting or verbal feedback~\citep{madaan2023self,shinn2023reflexion}. Without external feedback, however, intrinsic self-correction does not consistently improve accuracy and can even degrade it~\citep{huang2024large}. Reinforcement learning can train corrective behavior directly~\citep{kumar2025training}, while verification and backtracking may also emerge from outcome-based training~\citep{Guo_2025,gandhi2025cognitive}, although their occurrence does not reveal whether an individual correction was warranted. Credit-assignment methods address this limitation with denser supervision, including process-level scores from human or rollout-derived signals~\citep{lightman2024let,wang2024math}, Monte Carlo estimates from intermediate states~\citep{kazemnejad2024vineppo}, and advantages defined over turns or comparable states~\citep{wei2025reinforcing,feng2026group}. More recent work estimates the effect of tool use or individual steps through before--after comparisons and controlled counterfactual interventions~\citep{feng2026taco,meng2026policy}, while token-level methods concentrate optimization on selected positions~\citep{wang2026beyond,jin2026entropy}. In contrast to assigning credit over generic steps, turns, or tokens, our setting exposes the commit-or-re-examine choice as an explicit decision at a known search state, allowing its alternatives to be evaluated directly and credit to be localized to the decision itself.

\section{The \resee{} Benchmark}
\label{sec:data}

\begin{figure}[tbh]
\centering
\includegraphics[width=\textwidth]{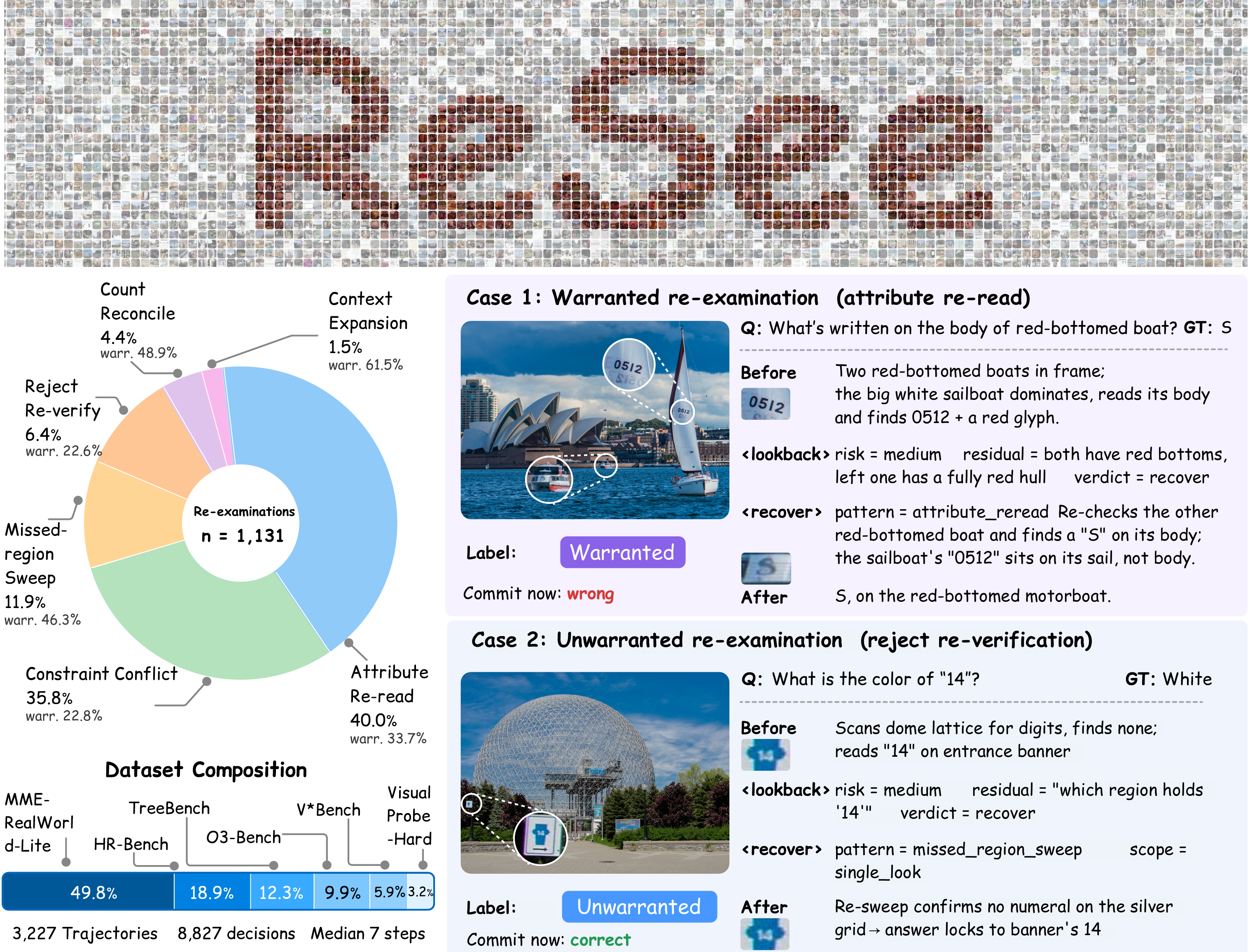}
\vspace{-10pt}
\caption{Overview of \resee{}. The left panel shows the sources and six re-examination patterns, with their warranted rates. The right panel contrasts a warranted and an unwarranted re-examination and shows the fields recorded at each decision.}
\vspace{-10pt}
\label{fig:resee}
\end{figure}

\paragraph{Task Formulation.}
\resee{} evaluates the commit-or-re-examine decision that recurs throughout long-horizon visual search. At a decision state \(h\), the model has accumulated a search history and formed a candidate answer, but may either commit or acquire another observation. We denote this decision by \(v \in \{\textsc{recover},\textsc{skip}\}\). The benchmark asks whether re-examination has positive state-dependent utility relative to immediate commitment:
\begin{equation}
\tau(h)=R_{\mathrm{rec}}(h)-R_{\mathrm{skip}}(h)-\lambda c(h),
\end{equation}
where \(R_{\mathrm{rec}}(h)\) and \(R_{\mathrm{skip}}(h)\) are the rewards reached by re-examining and by committing at \(h\), and \(\lambda c(h)\) is the priced cost of the additional observation. Re-examination at \(h\) is warranted when \(\tau(h)>0\), and immediate commitment is preferable otherwise. Unlike evaluation by final answers, this directly measures whether more evidence is warranted at the decision state.

\paragraph{Benchmark Construction.}
\resee{} contains \(3{,}227\) visual-search trajectories drawn from six high-resolution benchmarks: V*~\citep{wu2024vstar}, HR-Bench~\citep{wang2025divide}, MME-RealWorld~\citep{zhang2025mme}, TreeBench~\citep{wang2026traceable}, O3-Bench~\citep{li2025insight}, and VisualProbe~\citep{lai2026mini}. The resulting questions span attribute recognition, spatial reasoning, text reading, relational reasoning, and counting, with the target evidence occupying a median of only \(0.8\%\) of a \(3.39\)-MP image. Across these trajectories, we record \(8{,}827\) explicit decisions, including \(1{,}131\) re-examinations and \(7{,}696\) commitments. Each state preserves the accumulated evidence, candidate answer, remaining uncertainty, and the model's proposed verdict, and each re-examination further specifies its target region and search operation. The resulting decisions cover six recurring re-examination patterns, summarized in Figure~\ref{fig:resee} and Appendix~\ref{app:stats}.

\paragraph{Counterfactual Adjudication.}
We counterfactually adjudicate \(1{,}862\) decision states by replaying the same state under both verdicts and comparing their downstream returns. Labels therefore follow from the consequence of the decision rather than the model's own verdict or a subjective judgment of whether it is warranted. Among \(883\) adjudicated re-examinations, \(279\) (\(31.6\%\)) carry positive utility, whereas \(918\) of \(979\) adjudicated commitments (\(93.8\%\)) are safe to retain. Both classes are required to measure selectivity, since a policy that always re-examines attains perfect recall on the positive class alone. \resee{} therefore evaluates not re-examination itself, but the ability to distinguish states in which further evidence is valuable from those in which commitment is preferable.

\section{Methodology}
\label{sec:method}

\paragraph{Problem Formulation.}
Consider a visual-search rollout $\zeta=(u_1,a_1,\ldots,u_T,a_T)$ generated by a policy $\pi$, where $u_t$ and $a_t$ denote the observation and action at turn $t$. At each decision state $h_j$, the model has accumulated a search history and formed a candidate answer, and chooses a verdict $v_j\in\{\textsc{recover},\textsc{skip}\}$, corresponding to acquiring another observation or committing with the current evidence. Let $R(\zeta)$ denote the trajectory reward and $Q^\pi(h_j,v)$ the expected return after taking verdict $v$ at $h_j$ and following $\pi$ thereafter. We define the state-dependent utility of re-examination as
\begin{equation}
\tau^\pi(h_j)
=
Q^\pi(h_j,\textsc{recover})
-
Q^\pi(h_j,\textsc{skip})
-
\lambda c(h_j),
\label{eq:utility}
\end{equation}
where $c(h_j)$ is the cost of the additional observation and $\lambda \ge 0$ sets its price. This policy-dependent quantity is the expected-return counterpart of the paired-outcome benchmark utility $\tau(h)$ defined in Section~\ref{sec:data}. Re-examination is warranted when $\tau^\pi(h_j)>0$ and immediate commitment is preferable otherwise. The objective is to preserve task-level performance while selecting re-examination only at states where its expected benefit exceeds its interaction cost. Since a rollout realizes only one verdict at each state, this utility is not directly available from trajectory-level supervision.

\begin{figure}[tbh]
\centering
\includegraphics[width=\textwidth]{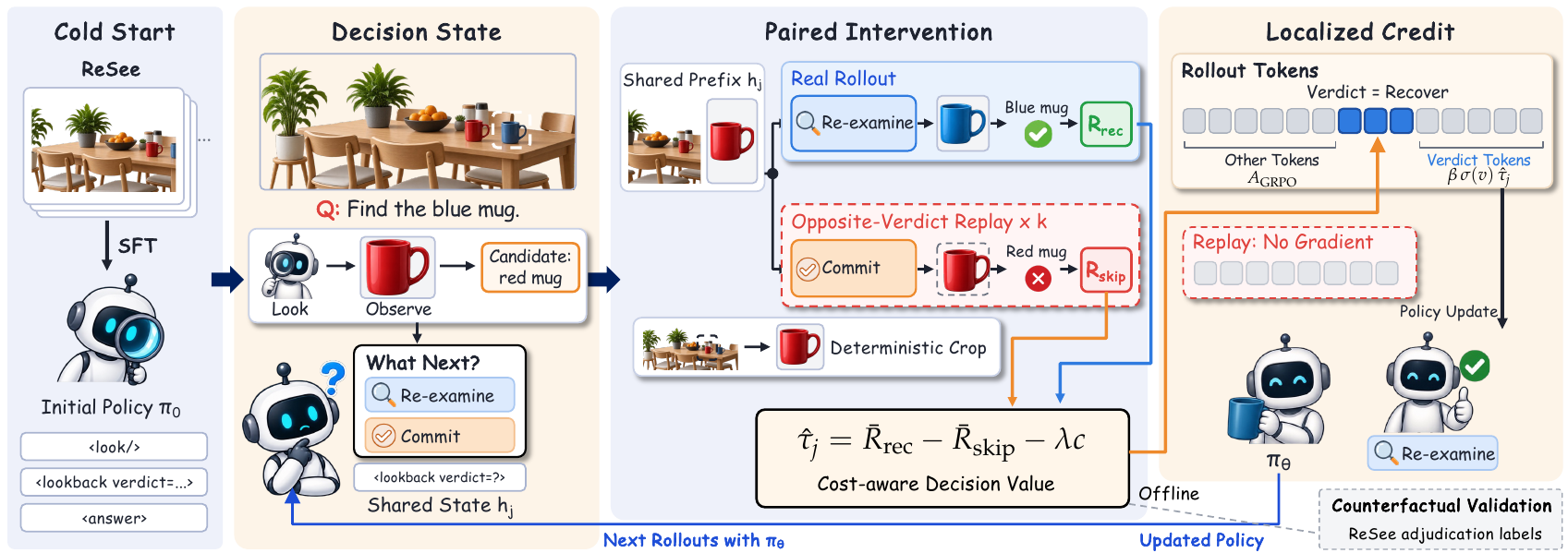}
\vspace{-8pt}
\caption{\care{}. Each on-policy verdict is paired with a replay under the opposite verdict from the same decision state. Their cost-adjusted return difference credits only the verdict tokens.}
\label{fig:method}
\vspace{-8pt}
\end{figure}

\paragraph{Overview.}
Figure~\ref{fig:method} summarizes \care{}. The curated \resee{} cold start establishes the visual-search protocol and an explicit commit-or-re-examine verdict. At up to $M$ decision states in each on-policy rollout, the observed continuation is paired with an opposite-verdict replay from the same state. Their cost-adjusted return difference estimates $\tau^\pi(h_j)$ and credits only the verdict tokens. Other tokens keep GRPO's group-relative advantage~\citep{shao2024deepseekmathpushinglimitsmathematical} (Algorithm~\ref{alg:care}).

\paragraph{Paired Counterfactual Estimation.}
At a decision state $h_j$, one verdict is taken by the on-policy rollout, while the alternative is imposed after resetting to the same environment state. This reset is exact because each visual observation is a deterministic crop of a fixed image: identical image--box pairs reproduce identical observations. The two branches therefore share the complete pre-decision context, including all preceding tokens and observations, while their post-verdict continuations are sampled independently from the current policy under identical decoding settings. Let $\bar R_{\mathrm{rec}}$ and $\bar R_{\mathrm{skip}}$ denote the empirical returns of the two branches at $h_j$, where the replayed branch is averaged over $k$ independent replays. We estimate the cost-aware value of re-examination as
\begin{equation}
\hat{\tau}_j =
\mathrm{clip}\!\left(
\bar R_{\mathrm{rec}}
-
\bar R_{\mathrm{skip}}
-
\lambda c(h_j),
-\kappa,\kappa
\right),
\label{eq:tau-hat}
\end{equation}
where $\kappa > 0$ bounds the estimate. Before clipping, this estimator is unbiased conditional on $h_j$ when both continuations follow their corresponding conditional policies, and increasing $k$ reduces the sampling variance of the replayed branch. Importantly, the observation cost enters only this decision contrast rather than the trajectory reward, so it changes the threshold for re-examination without penalizing unrelated search or reasoning tokens.

\paragraph{Localized Credit Assignment.}
The commit-or-re-examine verdict occupies an explicit span in the model output, so its token positions can be identified directly from the interaction protocol rather than inferred from entropy, attention, or state similarity. For the $i$-th rollout $\zeta_i$ of a sampling group, let $R_i = R(\zeta_i)$ be its reward, $v_{i,j}$ and $\hat\tau_{i,j}$ the verdict and estimate at its $j$-th decision state, and $\mathcal{V}_{i,j}$ the positions of that verdict's tokens. Define the sign $s(\textsc{recover})=+1$ and $s(\textsc{skip})=-1$. We replace the trajectory-level advantage on these positions with the signed counterfactual value,
\begin{equation}
\tilde A_{i,t} =
\begin{cases}
\beta\,s(v_{i,j})\,\hat{\tau}_{i,j},
& t \in \mathcal{V}_{i,j}, \\[3pt]
\dfrac{R_i-\mu_{g(i)}}{\sigma_{g(i)}+\delta},
& \text{otherwise},
\end{cases}
\label{eq:adv}
\end{equation}
where $\beta > 0$ scales the counterfactual credit, $\mu_{g(i)}$ and $\sigma_{g(i)}$ are the mean and standard deviation of the rewards in the sampling group $g(i)$, and $\delta$ is a small constant. Thus, a positive $\hat{\tau}_{i,j}$ reinforces \textsc{recover} and suppresses \textsc{skip}, whereas a negative value induces the opposite update, and the counterfactual credit is isolated to the decision that controls whether search continues.

\paragraph{Policy Optimization Objective.}
We optimize the resulting token-level advantages with the clipped policy-gradient objective used by GRPO. To preserve support for both verdicts during training, an entropy bonus is applied only to the verdict positions $\mathcal V = \bigcup_{i,j}\mathcal V_{i,j}$:
\begin{equation}
\mathcal{L}(\theta)
=
-\mathbb{E}_{t}\!\left[
\min\!\left(
r_t(\theta)\tilde A_t,\,
\mathrm{clip}\!\left(
r_t(\theta),
1-\varepsilon_{\mathrm{lo}},
1+\varepsilon_{\mathrm{hi}}
\right)\tilde A_t
\right)
\right]
+
\gamma\,\mathbb{E}_{t}\!\left[\mathrm{KL}_t\right]
-
\eta\,\mathbb{E}_{t\in\mathcal V}[H_t],
\label{eq:loss}
\end{equation}
where $y_t$ is the $t$-th output token, $r_t(\theta)=\pi_\theta(y_t\mid y_{<t})/\pi_{\mathrm{old}}(y_t\mid y_{<t})$, and $1-\varepsilon_{\mathrm{lo}}$ and $1+\varepsilon_{\mathrm{hi}}$ are the lower and upper clipping bounds. $\mathrm{KL}_t$ is the token-level divergence from the cold-start policy weighted by $\gamma$, and $H_t$ is the policy entropy weighted by $\eta$ and applied only to verdict tokens. Replays receive no gradient, and adjudication labels are used only to evaluate $\hat{\tau}$ offline.

\begin{proposition}[Verdict-Contrast Alignment]\label{prop:decision-alignment}
Fix a search state $h$ under a binary verdict choice, with $p=\pi(\textsc{recover}\mid h)\in(0,1)$ and $z=\log[p/(1-p)]$. Hold the conditional continuation policies fixed. Assume exact reset, independent finite-variance returns, and opposite-verdict replays from the corresponding conditional policy. For the unclipped paired contrast $D_v$ and $k\geq1$ replays, the expected verdict score gradient is
\begin{equation}
\mathbb E\!\left[\beta s(v)D_v\,\partial_z\log\pi(v\mid h)\mid h\right]
=2\beta p(1-p)\tau^\pi(h).
\label{eq:verdict-alignment}
\end{equation}
The contrast has conditional variance $\sigma_v^2+\sigma_{\neg v}^2/k$, where $\sigma_v^2$ is the return variance under $v$. Clipping preserves the expected gradient's sign for $|\tau^\pi(h)|>B_\kappa(h)$. Appendix~\ref{app:alignment-proof} characterizes this distortion and derives a finite-replay bound on the probability of a clipped verdict-score sign error.
\end{proposition}

\section{Experiments}
\label{sec:exp}

\paragraph{Benchmarks and Metrics.} We evaluate on our $980$-question split drawn from six source benchmarks~\citep{wu2024vstar,wang2026traceable,lai2026mini,wang2025divide,zhang2025mme,li2025insight}: $587$ annotated records and a $393$-record zero-shot set unseen in training. Accuracy is Avg@$32$ under a common scorer, reported as the macro average over the six sources so that each contributes equally. These results use our split and scoring protocol. The separate $468$-question uniform subset is the comparison set for published benchmark numbers. Against adjudication labels, we report recall, precision and the over-examination rate $P(\text{re-examine} \mid \text{unwarranted})$ for the re-examination decision. Appendix~\ref{app:setup} gives the grading rules and full metric definitions.

\paragraph{Baselines.}
Under the \resee{} tool protocol, we benchmark eight instruction-tuned VLMs and five released visual-search agents: Pixel Reasoner~\citep{su2026pixel}, DeepEyes~\citep{zheng2026deepeyes}, ViGoRL~\citep{sarch2026grounded}, TreeVGR~\citep{wang2026traceable}, and Mini-o3~\citep{lai2026mini}. We also include their common Qwen2.5-VL-7B backbone~\citep{bai2025qwen25vltechnicalreport}. We further evaluate state-agnostic controls that instruct the policy to re-examine, commit, or sample a random verdict, scoring the actions actually taken. For learning-based comparisons, \textsc{Zero-shot} uses the backbone without task-specific prompting or training, \textsc{Prompted} adds a re-examination instruction without parameter updates, and \textsc{SFT} stops after the common \resee{} cold start. All RL methods follow the same protocol: GRPO~\citep{shao2024deepseekmathpushinglimitsmathematical} uses trajectory-level advantages, GiGPO~\citep{feng2026group} adds step-level credit from repeated states, and Entropy~\citep{wang2026beyond} restricts updates to high-entropy tokens. GRPO-Pruned removes counterfactually unwarranted cold-start demonstrations to isolate data filtering from decision-level credit assignment. More details are given in Appendix~\ref{app:baselines}.

\paragraph{Implementation Details.} The two backbones are Qwen3-VL-8B-Instruct~\citep{bai2025qwen3vltechnicalreport} and InternVL3-8B~\citep{zhu2025internvl3exploringadvancedtraining}, with Qwen3-VL-8B used for framework analyses. Cold starts use the six variants of Appendix~\ref{app:care} and are full-parameter SFT with the vision tower and projector frozen, ZeRO-3 with optimizer offload~\citep{rajbhandari2020zero,ren2021zero}, learning rate $1{\times}10^{-5}$, three epochs, and context length $32{,}768$. RL uses GRPO \citep{shao2024deepseekmathpushinglimitsmathematical} with learning rate $1{\times}10^{-6}$, group size $8$, batch size $128$, sampling temperature $1.0$, one epoch over $11{,}500$ prompts, clipping $(\varepsilon_{\mathrm{lo}}, \varepsilon_{\mathrm{hi}}) = (0.2, 0.3)$, a token-level KL weight $\gamma = 10^{-3}$, and at most six generation rounds per trajectory. Both stages share a pixel budget of $[4{\times}10^{4},\,2{\times}10^{6}]$ per image. \care{} uses $k=1$ counterfactual replay per decision, at most $M=3$ decisions per trajectory, $\beta = 1$, clip $\kappa = 1$, an observation cost weight $\lambda = 2{\times}10^{-4}$, and an entropy weight $\eta = 10^{-2}$ on the verdict span.

\subsection{Main Results}
\label{sec:bench-results}

\paragraph{Benchmarking on \resee{}.}
Table~\ref{tab:bench} shows that answer accuracy alone does not predict whether a model re-examines at the right decision states. Several instruction-tuned models remain near the $50\%$ balanced-accuracy chance level under the CoT prompt, despite strong answer accuracy on the benchmark sources. Prompts that favor re-examination increase recall but also increase over-examination, while some released search agents attain higher precision by making few requests and leave warranted states unexamined. The resulting gap between answer quality and decision quality supports evaluating the verdict against its downstream consequence.
\begin{table}[tbh]
\centering
\vspace{-12pt}
\caption{Benchmarking on \resee{}. Balanced accuracy is shaded relative to the $50\%$ chance level.}
\label{tab:bench}
\setlength{\tabcolsep}{3.6pt}
\resizebox{\textwidth}{!}{%
\begin{tabular}{l ccccc c cccc c}
\toprule
 & \multicolumn{6}{c}{\textbf{Answer accuracy} $\uparrow$} & \multicolumn{4}{c}{\textbf{Re-examination decision}} & \textbf{Cost} \\
\cmidrule(lr){2-7}\cmidrule(lr){8-11}\cmidrule(lr){12-12}
\textbf{Model} & V* & HR-Bench & MME-RW & TreeBench & O3-Bench & \textbf{Macro} & Recall & Prec. & Over-ex.\,$\downarrow$ & Bal.\ acc. & Looks\,$\downarrow$ \\
\rowcolor{ReseeBand}\multicolumn{12}{l}{\textit{Instruction-tuned vision--language models}} \\
Qwen3.8-27B &\textbf{92.6}&\textbf{88.4}&48.8&\textbf{61.6}&\textbf{59.1}&\textbf{70.1}&4.4&14.3&5.9&\cellcolor{ReseeRed!10}49.3&3.4\\
DeepSeek-V4-Flash &79.5&\underline{80.0}&53.9&\underline{54.0}&36.6&\underline{60.8}&4.4&11.1&7.7&\cellcolor{ReseeRed!14}48.4&25.4\\
Qwen3-VL-235B-A22B &\underline{81.6}&77.6&43.6&49.5&35.9&57.6&\underline{31.4}&18.0&31.9&\cellcolor{ReseeRed!9}49.7&5.0\\
Muse-Glimmer-30B &55.3&69.1&47.8&44.5&\underline{40.9}&51.5&0.0&0.0&1.5&\cellcolor{ReseeRed!10}49.3&8.2\\
Qwen3-VL-32B &77.4&74.0&44.7&47.0&26.2&53.9&\textbf{31.6}&\underline{33.3}&18.2&\cellcolor{ReseeBlue!34}\textbf{56.7}&0.8\\
MiMo-V2.5 &69.5&78.4&43.8&37.9&21.9&50.3&6.2&\underline{33.3}&\underline{1.2}&\cellcolor{ReseeBlue!18}52.5&0.3\\
GLM-4.6V &37.4&7.4&20.4&10.3&5.6&16.2&30.6&26.8&19.2&\cellcolor{ReseeBlue!30}\underline{55.7}&3.9\\
Grok-4.6 &7.9&9.7&4.4&2.8&1.6&5.3&4.0&25.0&2.8&\cellcolor{ReseeBlue!10}50.6&5.1\\
\rowcolor{ReseeBand}\multicolumn{12}{l}{\textit{Released visual search agents}} \\
TreeVGR-7B~\citep{wang2026traceable} &70.2&57.9&\textbf{57.9}&42.7&33.1&52.4&18.8&18.0&18.6&\cellcolor{ReseeBlue!8}50.1&\textbf{0.01}\\
Mini-o3-7B~\citep{lai2026mini} &73.8&57.1&\underline{54.6}&43.6&32.4&52.3&12.5&24.6&8.7&\cellcolor{ReseeBlue!15}51.9&4.22\\
DeepEyes-7B~\citep{zheng2026deepeyes} &63.7&54.2&51.1&35.7&27.6&46.5&2.9&\textbf{43.8}&\textbf{0.5}&\cellcolor{ReseeBlue!12}51.2&0.27\\
Pixel-Reasoner-7B~\citep{su2026pixel} &64.9&53.4&46.2&39.0&25.4&45.8&6.2&22.0&4.6&\cellcolor{ReseeBlue!11}50.8&0.40\\
Qwen2.5-VL-7B~\citep{bai2025qwen25vltechnicalreport} &61.3&53.4&43.6&34.5&24.3&43.4&1.9&22.5&1.3&\cellcolor{ReseeBlue!9}50.3&\underline{0.22}\\
ViGoRL-7B~\citep{sarch2026grounded} &61.9&50.6&44.2&33.8&22.4&42.6&7.7&13.0&10.9&\cellcolor{ReseeRed!14}48.4&4.24\\
\bottomrule
\end{tabular}}
\vspace{-12pt}
\end{table}

\paragraph{Learning on \resee{}.} Table~\ref{tab:main} shows that \care{} achieves the highest macro accuracy and balanced accuracy on both backbones while using fewer looks than GRPO. On Qwen3-VL, it gains $2.5$ macro points and $11.1$ balanced-accuracy points, with the same direction of improvement on InternVL3. The answer-accuracy gains are larger on O3-Bench and VisualProbe. These gains coincide with higher recall of warranted requests and lower over-examination. \care{} also improves on GRPO-Pruned, indicating that removing unwarranted demonstrations alone does not account for the full result. Together, the two backbones support verdict-level credit as a way to improve both answer quality and selective visual re-examination.
\begin{table}[tbh]
\centering
\vspace{-8pt}
\caption{Main results on our six-source split. Macro subscripts report run-to-run standard deviations.}
\label{tab:main}
\resizebox{\textwidth}{!}{%
\begin{tabular}{l cccccc c cccc c}
\toprule
 & \multicolumn{7}{c}{\textbf{Answer accuracy} $\uparrow$} & \multicolumn{4}{c}{\textbf{Re-examination decision}} & \textbf{Cost} \\[1pt]
\cmidrule(lr){2-8}\cmidrule(lr){9-12}\cmidrule(lr){13-13}
\textbf{Method} & V* & HR-Bench & MME-RW & TreeBench & O3-Bench & VisualProbe & \textbf{Macro} & Recall & Prec. & Over-ex.\,$\downarrow$ & Bal.\ acc. & Looks\,$\downarrow$ \\
\midrule
\multicolumn{13}{l}{\textbf{Qwen3-VL-8B}} \\
Always-prompt &53.8&57.6&46.5&32.4&23.6&19.7&38.9$_{\pm 1.8}$&\textbf{66.5}&17.3&70.0&48.2&2.16\\
Never-prompt &56.0&58.8&45.7&31.6&21.1&\underline{22.9}&39.3$_{\pm 2.0}$&0.7&24.2&\underline{0.4}&50.1&2.19\\
Random verdict &53.7&57.7&47.0&32.1&22.2&\textbf{23.6}&39.4$_{\pm 1.4}$&17.5&17.3&18.4&49.5&2.20\\
\cmidrule(l{2pt}r{2pt}){1-13}
Zero-shot &\textbf{68.5}&56.9&23.4&\textbf{41.1}&29.6&0.8&36.7$_{\pm 1.1}$&0.0& -- &\textbf{0.0}&50.0&\textbf{0.00}\\
Prompted &46.9&50.6&34.1&33.4&11.5&17.2&32.3$_{\pm 1.5}$&11.1&22.9&8.2&51.5&3.53\\
SFT &55.7&59.3&48.5&33.2&23.4&15.5&39.3$_{\pm 1.8}$&30.4&22.1&23.5&53.5&2.14\\
\cmidrule(l{2pt}r{2pt}){1-13}
GRPO &65.8&67.1&58.2&36.8&31.5&16.8&46.0$_{\pm 1.6}$&23.1&27.3&13.6&54.8&2.09\\
GiGPO &65.9&\underline{67.4}&58.3&37.0&31.8&17.4&46.3$_{\pm 1.6}$&23.1&26.7&14.0&54.5&2.15\\
Entropy &65.5&67.0&58.1&37.1&32.0&16.9&46.1$_{\pm 1.7}$&21.2&28.2&11.9&54.6&2.06\\
GRPO-Pruned &65.1&67.2&\underline{58.7}&37.4&\underline{32.9}&18.1&\underline{46.6$_{\pm 1.5}$}&25.0&\underline{37.1}&9.3&\underline{57.8}&1.99\\
\cmidrule(l{2pt}r{2pt}){1-13}
\rowcolor{ReseeBlue!10}\textbf{\care{} (Ours)} &\underline{66.8}&\textbf{68.1}&\textbf{59.9}&\underline{39.0}&\textbf{35.1}&21.9&\textbf{48.5$_{\pm 1.3}$}&\underline{42.3}&\textbf{46.8}&10.6&\textbf{65.9}&\underline{1.97}\\
\midrule
\multicolumn{13}{l}{\textbf{InternVL3-8B}} \\
Always-prompt &51.5&57.7&50.4&37.6&12.7&21.2&38.5$_{\pm 1.5}$&\textbf{54.7}&21.0&45.4&54.7&2.54\\
Never-prompt &53.2&58.4&49.7&36.8&14.9&22.6&39.3$_{\pm 1.7}$&1.9&20.0&1.7&50.1&2.50\\
Random verdict &52.4&58.0&50.2&37.2&14.2&22.0&39.0$_{\pm 1.5}$&19.2&17.9&19.5&49.9&2.51\\
\cmidrule(l{2pt}r{2pt}){1-13}
Zero-shot &54.8&54.8&32.1&35.3&\textbf{35.6}&23.8&39.4$_{\pm 0.2}$&0.0& -- &\textbf{0.0}&50.0&\textbf{0.00}\\
Prompted &57.1&58.3&34.9&31.7&\underline{29.4}&15.5&37.8$_{\pm 0.2}$&5.3&\textbf{46.6}&\underline{1.5}&51.9&\underline{1.06}\\
SFT &53.0&58.7&51.0&38.1&15.4&21.7&39.7$_{\pm 1.6}$&32.7&20.7&27.5&52.6&2.49\\
\cmidrule(l{2pt}r{2pt}){1-13}
GRPO &58.9&62.5&55.3&41.8&23.8&25.2&44.6$_{\pm 1.5}$&26.9&26.9&16.1&55.4&2.34\\
GiGPO &\underline{59.2}&\underline{62.9}&55.2&\underline{42.4}&24.5&\underline{25.9}&\underline{45.0$_{\pm 1.5}$}&28.8&28.8&15.7&56.6&2.32\\
Entropy &58.8&62.8&55.6&42.2&24.1&25.3&44.8$_{\pm 1.6}$&25.0&28.9&13.6&55.7&2.27\\
GRPO-Pruned &59.0&62.7&\underline{55.8}&42.0&25.0&25.7&\underline{45.0$_{\pm 1.4}$}&28.8&34.9&11.9&\underline{58.5}&2.20\\
\cmidrule(l{2pt}r{2pt}){1-13}
\rowcolor{ReseeBlue!10}\textbf{\care{} (Ours)} &\textbf{59.8}&\textbf{63.6}&\textbf{57.0}&\textbf{44.1}&27.6&\textbf{29.3}&\textbf{46.9$_{\pm 1.2}$}&\underline{44.2}&\underline{43.4}&12.7&\textbf{65.8}&2.15\\
\bottomrule
\end{tabular}}
\vspace{-6pt}
\end{table}

\subsection{Framework Analysis}
\paragraph{Ablation of Decision-Level Credit.} Table~\ref{tab:ablation} shows that the paired verdict contrast and its localization are central to \care{}. Removing the verdict flip lowers macro accuracy by $4.3$ points, the largest ablation effect. Its recall also falls most sharply, consistent with paired comparison helping identify states that warrant another look. Without observation pricing, over-examination nearly doubles, while trajectory-level pricing restores selectivity only partly and diffuse credit reduces recall and precision. This pattern supports localizing credit and cost to the verdict under evaluation.

\begin{table}[tbh]
\centering
\vspace{-6pt}
\caption{Ablation of \care{}. Macro subscripts give run-to-run standard deviations. Red subscripts show relative degradation from the full method, with increases for Over-ex. and Looks.}
\label{tab:ablation}
\newcommand{\abdelta}[2]{\(#1_{\textcolor{ReseeRed}{\scriptstyle #2\%}}\)}
\resizebox{.985\textwidth}{!}{%
\begin{tabular}{l c cccc c}
\toprule
\textbf{Variant} & \textbf{Macro} $\uparrow$ & Recall $\uparrow$ & Prec.\ $\uparrow$ & Over-ex.\ $\downarrow$ & Bal.\ acc.\ $\uparrow$ & Looks $\downarrow$ \\
\midrule
\rowcolor{ReseeBlue!10}\textbf{\care{}} &\textbf{48.5$_{\pm 1.3}$}&\textbf{42.3}&\textbf{46.8}&\textbf{10.6}&\textbf{65.9}&\textbf{1.97}\\
\midrule
\emph{w/o} localization &45.8$_{\pm 1.7}$&\abdelta{30.8}{-27.2}&\abdelta{34.0}{-27.4}&\abdelta{13.1}{+23.6}&\abdelta{58.8}{-10.8}&\abdelta{2.08}{+5.6}\\
\emph{w/o} verdict flip &44.2$_{\pm 1.9}$&\abdelta{21.2}{-49.9}&\abdelta{25.6}{-45.3}&\abdelta{13.6}{+28.3}&\abdelta{53.8}{-18.4}&\abdelta{2.12}{+7.6}\\
\emph{w/o} cost &45.4$_{\pm 1.6}$&\abdelta{29.3}{-30.7}&\abdelta{24.4}{-47.9}&\abdelta{19.9}{+87.7}&\abdelta{54.7}{-17.0}&\abdelta{2.14}{+8.6}\\
w/ trajectory cost &46.5$_{\pm 1.6}$&\abdelta{\underline{34.6}}{-18.2}&\abdelta{38.3}{-18.2}&\abdelta{12.3}{+16.0}&\abdelta{61.2}{-7.1}&\abdelta{2.07}{+5.1}\\
\emph{w/o} verdict entropy &\underline{47.2$_{\pm 1.4}$}&\abdelta{\underline{34.6}}{-18.2}&\abdelta{\underline{40.0}}{-14.5}&\abdelta{\underline{11.4}}{+7.5}&\abdelta{\underline{61.6}}{-6.5}&\abdelta{\underline{2.05}}{+4.1}\\
\bottomrule
\end{tabular}
}%
\vspace{-11pt}
\end{table}

\begin{figure}[tbh]
\centering
\includegraphics[width=\textwidth]{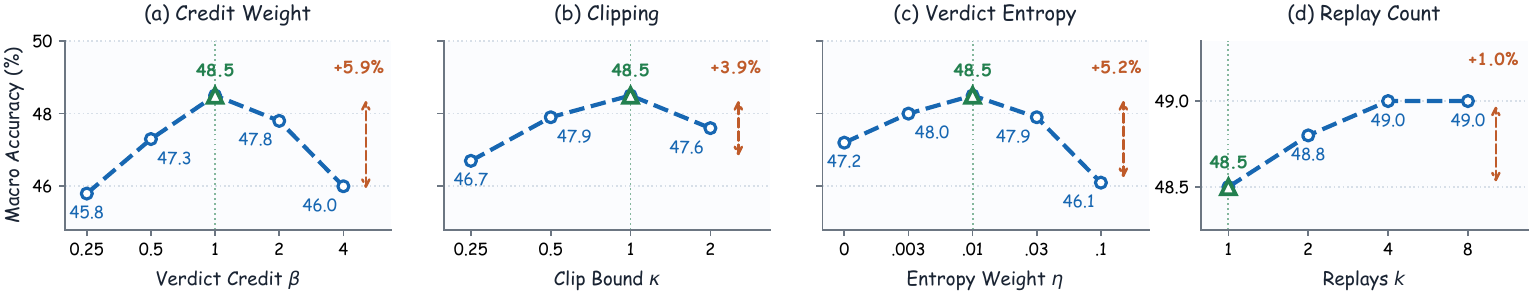}
\vspace{-18pt}
\caption{Qwen3-VL-8B macro accuracy across four parameter sweeps. Green triangles mark the main setting, and arrows report the percentage gain between the extremes.}
\label{fig:sensitivity}
\vspace{-9pt}
\end{figure}

\paragraph{Hyperparameter Sensitivity.}
Figure~\ref{fig:sensitivity} shows broad performance peaks for verdict credit and entropy, with a flatter response to clipping. Reducing $\beta$ to $0.25$ lowers macro accuracy by $2.7$ points, while balanced accuracy falls and visual requests rise. Increasing $\beta$ beyond its reference value also reduces answer and decision quality, giving the sweep a clear interior optimum. Entropy follows a similar pattern, whereas clipping causes smaller variation across its tested range. Additional replays yield diminishing accuracy gains: macro accuracy rises from $48.5$ at $k=1$ to $49.0$ at $k=4$ and remains there at $k=8$. The curves indicate that verdict-credit calibration matters more than fine adjustments to clipping, while a small replay budget captures most of the observed gain.

\begin{wrapfigure}{r}{0.41\textwidth}
\vspace{-7pt}
\centering
\includegraphics[width=\linewidth]{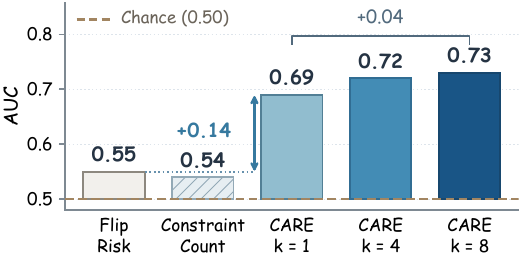}
\vspace{-18pt}
\caption{Utility-ranking AUC.}
\label{fig:utility}
\vspace{-5pt}
\end{wrapfigure}

\paragraph{Counterfactual Utility.} Counterfactual utility tests whether paired-rollout return differences rank decisions by the benefit of re-examination. Figure~\ref{fig:utility} shows that the single-replay estimator reaches $0.69$ AUC, compared with $0.55$ for self-assessed answer-flip risk and $0.54$ for unresolved-constraint count. The paired estimate therefore ranks warranted decisions more reliably than either annotator-recorded signal. Additional replays raise AUC to $0.73$ at $k=8$, with smaller gains beyond the single-replay setting. This pattern supports estimating the consequence of the verdict rather than relying on stated uncertainty alone.

\begin{wrapfigure}{r}{0.42\textwidth}
\vspace{-14pt}
\centering
\includegraphics[width=\linewidth]{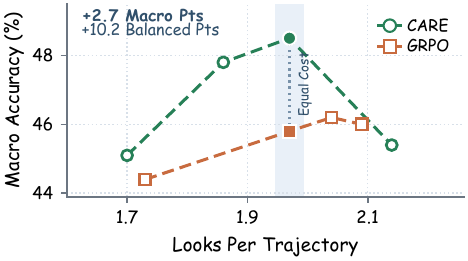}
\vspace{-18pt}
\caption{Macro accuracy and visual cost for \care{} and GRPO.}
\label{fig:cost}
\vspace{-12pt}
\end{wrapfigure}

\paragraph{Decision Quality at Equal Visual Cost.} Figure~\ref{fig:cost} shows that \care{} retains higher accuracy and selectivity than cost-tuned GRPO at the same mean number of visual requests. At $1.97$ looks per trajectory, it leads by $2.7$ macro points and $10.2$ balanced-accuracy points. Across the nonzero prices, increasing observation cost reduces requests for both policies, while \care{} retains higher balanced accuracy. At the strongest price, both policies lose answer accuracy as looks fall, consistent with useful evidence being suppressed. The gap at comparable cost indicates that request budgeting alone does not explain the decision-quality gains, supporting the role of assigning observation cost and paired return contrast directly to each verdict.

\paragraph{Out-of-Distribution Generalization.} To assess transfer beyond the \resee{} training split, we evaluate six test sets in high-resolution perception and extended visual search under their official scorers (Appendix~\ref{app:ood-benchmarks}). Table~\ref{tab:ood} shows that \care{} scores higher than GRPO with fewer visual requests on all six sets. The largest gains are on VistaHop and VisualNeedle, at $4.0$ and $3.6$ points, respectively. Gains on remote-sensing and multi-image tasks extend this pattern across visual-search settings.
\begin{table}[tbh]
\centering
\caption{OOD results for \care{} with Qwen3-VL-8B-Instruct. Obs. reports mean visual tool calls, and Gain is the relative score improvement over GRPO.}
\label{tab:ood}
\setlength{\tabcolsep}{4pt}
\footnotesize
\resizebox{\textwidth}{!}{%
\begin{tabular}{l c cc cc cc ccc}
\toprule
 & & \multicolumn{2}{c}{\textbf{Zero-shot}} & \multicolumn{2}{c}{\textbf{SFT}} & \multicolumn{2}{c}{\textbf{GRPO}} & \multicolumn{3}{c}{\textbf{\care{}}} \\
\cmidrule(lr){3-4}\cmidrule(lr){5-6}\cmidrule(lr){7-8}\cmidrule(lr){9-11}
\textbf{Benchmark} & \textbf{Metric} & Score\,$\uparrow$ & Obs.\,$\downarrow$ & Score\,$\uparrow$ & Obs.\,$\downarrow$ & Score\,$\uparrow$ & Obs.\,$\downarrow$ & Score\,$\uparrow$ & Obs.\,$\downarrow$ & Gain\,(\%) \\
\midrule
\rowcolor{ReseeBand}\multicolumn{11}{l}{\textit{High-Resolution Perception}} \\
HR-Bench 8K~\citep{wang2025divide} & Acc. &66.5&0.00&68.0&2.35&\underline{69.5}&2.24&\textbf{71.0}&\underline{2.11}&+2.2\\
MagnifierBench~\citep{li2023otterhd} & Acc. &52.3&0.00&56.2&2.56&\underline{58.7}&2.41&\textbf{61.5}&\underline{2.25}&+4.8\\
XLRS-Bench-Lite~\citep{wang2025xlrs} & Acc. &42.4&0.00&44.1&3.02&\underline{46.0}&2.90&\textbf{48.3}&\underline{2.72}&+5.0\\
\rowcolor{ReseeBand}\multicolumn{11}{l}{\textit{Extended Visual Search}} \\
VistaHop~\citep{he2026vistahop} & Pass@1 &8.3&0.00&11.7&5.64&\underline{14.0}&5.52&\textbf{18.0}&\underline{5.18}&+28.6\\
VisualNeedle~\citep{chen2026visualneedle} & Acc. &17.0&0.00&36.3&3.10&\underline{40.7}&2.94&\textbf{44.3}&\underline{2.70}&+8.8\\
Visual Haystacks-SN-50~\citep{wu2025visual} & Acc. &50.7&0.00&52.6&4.84&\underline{54.0}&4.58&\textbf{55.8}&\underline{4.26}&+3.3\\
\bottomrule
\end{tabular}}
\vspace{-16pt}
\end{table}

\begin{figure}[tbh]
\centering
\includegraphics[width=1\textwidth]{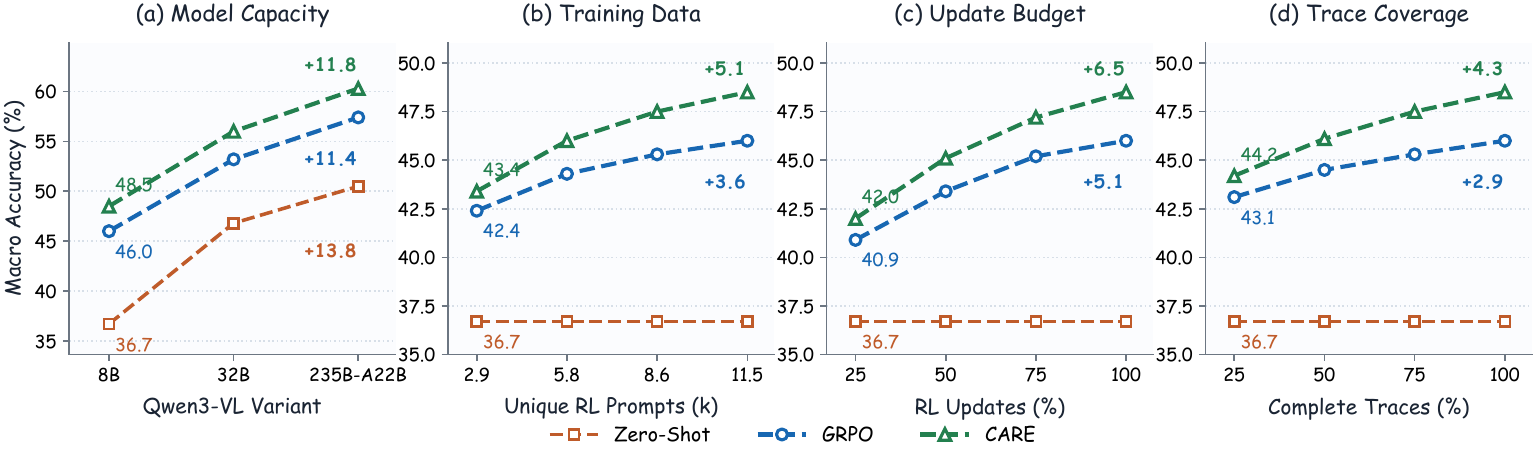}
\vspace{-14pt}
\caption{Six-source macro Avg@$32$ under the evaluation protocol of Table~\ref{tab:main}. Model scaling includes 235B-A22B sparse experts, and other sweeps use 8B.}
\label{fig:scaling}
\vspace{-8pt}
\end{figure}

\paragraph{Scaling Behaviours.} We evaluate four scaling dimensions: model capacity, training data, update budget, and complete-trace coverage. Figure~\ref{fig:scaling} shows that \care{} retains its advantage from 8B through 235B-A22B and across the three training sweeps. Its macro-accuracy lead reaches $2.9$ points on the largest variant, while extra prompts and complete traces yield steeper gains than for GRPO. These trends suggest that decision-level credit remains useful at larger scales.

\begin{wrapfigure}{R}{0.62\textwidth}
\vspace{-12pt}
\centering
\includegraphics[width=\linewidth]{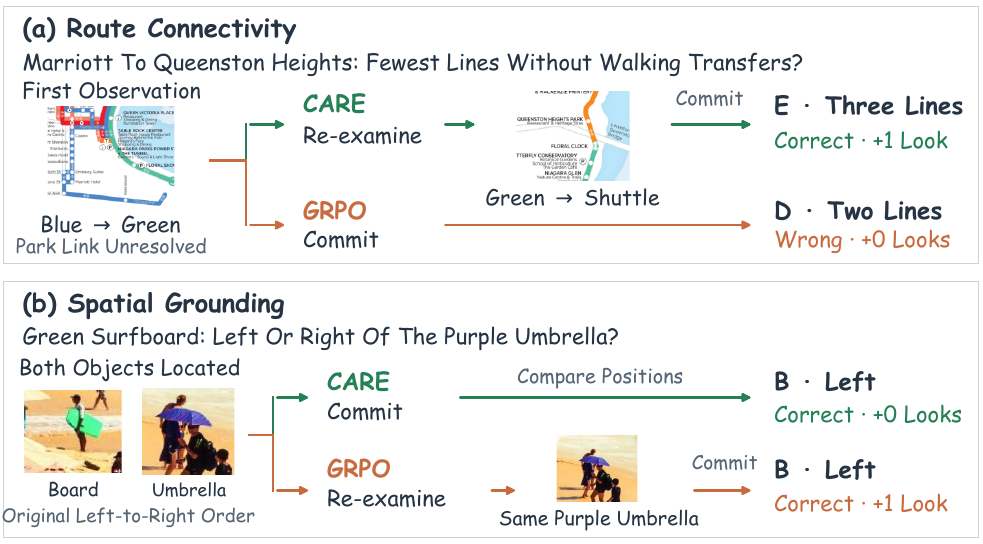}
\vspace{-14pt}
\caption{\care{} and GRPO trajectories. Extra looks start at the shared state. Full cases appear in Appendix~\ref{app:case-full}.}
\label{fig:case-main}
\vspace{-10pt}
\end{wrapfigure}

\paragraph{Case Studies.} Figure~\ref{fig:case-main} contrasts useful re-examination with timely commitment. On the route map, \care{} checks the missing connection before answering, whereas GRPO commits after the first transfer. In the beach scene, both objects are located and their spatial relation is settled. \care{} stops, while GRPO unnecessarily repeats the umbrella inspection. Across map connectivity and object relations, the contrast shows a shared decision criterion: re-examine when required evidence is missing and commit once it is resolved.

\section{Conclusion}
\label{sec:conclusion}
This paper proposes a framework for evaluating and learning selective re-examination in visual search. \resee{} provides explicit commit-or-re-examine decisions and counterfactual adjudication of their utility, supporting both benchmarking and policy learning. \care{} compares paired continuations from the same search state and assigns their cost-aware return difference to the verdict tokens. Extensive experiments on real-world visual-search tasks across two VLM backbones demonstrate that \care{} outperforms GRPO in accuracy and decision selectivity while reducing visual requests, with accuracy gains extending to six out-of-distribution benchmarks.

\subsection*{AI use statement}
In this work, we used LLM tools to polish wording and grammar, generate the icons in Figure~\ref{fig:intro}, and revise code. We also used a vision-language model to generate trajectories for the \resee{} benchmark. We carefully reviewed all AI-assisted content and take responsibility for the final content.

\subsection*{Reproducibility statement}
Section~\ref{sec:exp} specifies the evaluation split, metrics, baselines, and training configuration. Appendix~\ref{app:resee} documents dataset construction and counterfactual adjudication. Appendix~\ref{app:care} gives the training algorithm and cold-start variants, while Appendix~\ref{app:setup} defines answer grading and decision metrics.

\bibliographystyle{iclr2027_conference}
\bibliography{refs}

\clearpage
\appendix
\paragraph{Roadmap.} Appendix~\ref{app:resee} details dataset construction and human evaluation, Appendix~\ref{app:care} gives the training procedure and alignment proofs, and Appendix~\ref{app:setup} specifies the experimental protocol. Appendices~\ref{app:vlms}, \ref{app:traj}, and \ref{app:results} provide released-model benchmarking, case studies, and additional results, respectively.

\section{Details of \resee{}}
\label{app:resee}

\subsection{Annotation Protocol}
\label{app:protocol}
\paragraph{Trajectory Generation.} Qwen3-VL-Plus~\citep{bai2025qwen3vltechnicalreport} writes each trajectory from the image, question and reference answer, which anchors the search to the target evidence. A trajectory contains at most twelve steps. A step parses the question, scans the image, inspects a region, compares candidates, re-examines an earlier region, confirms the answer, or flags uncertainty. Every step that reads the image records its inspected box.

\paragraph{Conservative Verdicts.} At each decision state, the annotator records a verdict, unresolved constraints, estimated answer-flip risk and estimated cost. The protocol calls for re-examination when a constraint remains unresolved, with at most three re-examinations per trajectory. All $1{,}131$ re-examinations record an unresolved constraint, while $7{,}505$ of the $7{,}696$ commitments record none. Counterfactual adjudication compares continuation outcomes to assign the labels. Table~\ref{tab:schema} lists the recorded fields and six re-examination patterns. Each re-examination also records the revisited step, region and effort.

\paragraph{Counterfactual Adjudication.} Decision labels compare re-examination and commitment from the same state using the utility $\tau(h)$ of Section~\ref{sec:data}. The demonstrated re-examination continuation reaches the reference answer in $99.3\%$ of trajectories. The commitment outcome truncates the trajectory at the decision state and requires an answer from the evidence gathered so far. With answer correctness as the reward and priced cost $\lambda c(h) \in (0,1)$, a successful re-examination has $\tau(h) > 0$ exactly when commitment produces an incorrect answer. The same comparison labels re-examinations as warranted or unwarranted and commitments as premature or safe.

\begin{table}[h]
\centering
\caption{The top block lists fields recorded at a decision state, and the bottom block describes the six re-examination patterns.}
\label{tab:schema}
\resizebox{\textwidth}{!}{%
\begin{tabular}{l l l}
\toprule
\rowcolor{ReseeBand}\multicolumn{3}{l}{\textit{Decision record}} \\
\textbf{Field} & \textbf{Values} & \textbf{Meaning} \\
\midrule
Verdict & recover, skip & whether the annotator proposes to re-examine before committing \\
Candidate step & step index & the step that produced the answer currently held \\
Residual constraints & free text, possibly empty & constraints of the question not yet verified on the image \\
Flip risk & low, medium, high & self-assessed risk that the answer would change ($7{,}538$ / $637$ / $652$) \\
Cost estimate & four integers and flags & extra steps, attention units, need to zoom again, distance to the revisited step \\
Rationale & free text & justification of the verdict \\
Label & warranted, unwarranted, safe, premature & outcome of the adjudication, present on $1{,}862$ decisions \\
\midrule
Target step and region & step index, box & where a re-examination returns to and what it looks at \\
Scope & single look, re-inspect, re-sweep & effort the re-examination may spend ($648$ / $437$ / $46$) \\
\bottomrule
\end{tabular}}

\vspace{6pt}
\resizebox{\textwidth}{!}{%
\begin{tabular}{l l r r}
\toprule
\rowcolor{ReseeBand}\multicolumn{4}{l}{\textit{Re-examination patterns}} \\
\textbf{Pattern} & \textbf{The re-examination \ldots} & \textbf{Count} & \textbf{Warranted (\%)} \\
\midrule
Attribute re-read & reads again an attribute, such as text or color, that was read under doubt & 452 & 33.7 \\
Constraint conflict & checks a candidate against a constraint that the evidence seems to contradict & 405 & 22.8 \\
Missed-region sweep & searches a region that the earlier search left unvisited & 135 & 46.3 \\
Reject re-verify & returns to a candidate that was rejected earlier & 72 & 22.6 \\
Count reconcile & recounts after partial counts disagree & 50 & 48.9 \\
Context expansion & widens the view around a candidate to recover its context & 17 & 61.5 \\
\bottomrule
\end{tabular}}
\end{table}

\subsection{Dataset Statistics}
\label{app:stats}
Table~\ref{tab:dataset} reports the composition of \resee{} by source. Images have a median area of $3.39$\,MP and reach $86.25$\,MP, and the median target box occupies $0.8\%$ of the frame. Each trajectory becomes a multi-turn episode in which every look and every re-examination ends the assistant turn. The requested region is cut at full resolution with a $10\%$ margin, bounded at $1024$ pixels on its longer side and returned as the next observation, giving $6{,}333$ observations over $2{,}197$ episodes. Two filters separate training from the evaluation split: the $459$ trajectories whose reasoning quotes the reference answer leave supervised training, and a perceptual comparison against the evaluation images removes the $397$ training records built on $207$ near-duplicate images. On the $883$ adjudicated re-examinations, self-assessed flip risk has an AUC of $0.545$, unresolved-constraint count has an AUC of $0.533$, and each cost estimate has an AUC no greater than $0.51$.

\begin{table}[tbh]
\centering
\caption{Composition of \resee{} by source benchmark.}
\label{tab:dataset}
\resizebox{\textwidth}{!}{%
\begin{tabular}{l rr rrr rrrr}
\toprule
 & & & \multicolumn{3}{c}{\textbf{Decisions}} & \multicolumn{4}{c}{\textbf{Adjudicated}} \\
\cmidrule(lr){4-6}\cmidrule(lr){7-10}
\textbf{Source} & Traj. & Images & All & Re-ex. & Commit & Warr. & Unwarr. & Safe & Premature \\
\midrule
MME-RealWorld & 1{,}606 & 1{,}422 & 4{,}380 & 588 & 3{,}792 & 144 & 300 & 207 & 6 \\
HR-Bench & 611 & 193 & 1{,}509 & 72 & 1{,}437 & 9 & 54 & 156 & 1 \\
TreeBench & 398 & 395 & 1{,}205 & 172 & 1{,}033 & 41 & 84 & 176 & 1 \\
O3-Bench & 320 & 197 & 982 & 199 & 783 & 55 & 112 & 139 & 9 \\
V* & 190 & 190 & 437 & 21 & 416 & 1 & 15 & 145 & 0 \\
VisualProbe & 102 & 102 & 314 & 79 & 235 & 29 & 39 & 95 & 44 \\
\midrule
\rowcolor{ReseeBand}\textbf{Total} & 3{,}227 & 2{,}499 & 8{,}827 & 1{,}131 & 7{,}696 & 279 & 604 & 918 & 61 \\
\bottomrule
\end{tabular}}
\end{table}

\subsection{Human Evaluation}
\label{app:human}
Three expert raters from our research group, who received no compensation, assess the five dimensions in Table~\ref{tab:rubric} on source-stratified trajectories, verdict-stratified decisions, and pattern-stratified re-examinations, including the three rarest patterns. The adjudication sample balances warranted and unwarranted decisions. Raters see the images, transcripts, and returned crops without adjudication labels, reference answers, or other ratings. Agreement is measured by Fleiss' $\kappa$ for binary dimensions~\citep{fleiss1971measuring} and Krippendorff's $\alpha$ for ordinal dimensions~\citep{krippendorff2004reliability}. Table~\ref{tab:human} reports high question and decision-state validity and $83.3\%$ agreement with adjudication labels, with inter-rater $\kappa=0.73$. Lower agreement on thought--action faithfulness suggests that trajectory wording provides a less stable supervision signal than outcome-based decision labels.

\newcommand{\rlab}[1]{{\color{black!48}\scshape #1}\hspace{0.55em}}
\newcommand{\rubitem}[6]{%
  \makebox[\linewidth]{{\color{#1}\scshape\bfseries #2}\hfill{\color{black!42}\scriptsize #3}}\par
  \vspace{2.5pt}\textit{``#4''}\par
  \vspace{3pt}{\hangindent=3.5em\hangafter=1\rlab{positive}#5\par}
  \vspace{1.5pt}{\hangindent=3.5em\hangafter=1\rlab{failure}#6\par}}
\begin{table}[h]
\centering
\caption{Rubric used by the \resee{} raters.}
\label{tab:rubric}
\footnotesize
\begin{tabular}{@{}p{\linewidth}@{}}
\toprule
\rubitem{TrAsk}{Question Validity}{trajectory, binary}
{Given the question and the reference region outlined on the full image, is the answer settled by that region alone?}
{The region holds the evidence that settles the question, and an attentive reader could answer from the image without further context.}
{The region is empty or off-target, several regions answer equally well, or the question admits more than one answer.} \\
\arrayrulecolor{black!15}\midrule\arrayrulecolor{black}
\rubitem{TrPlan}{Decision-State Validity}{decision, binary}
{The search has returned some evidence. Has a candidate answer formed here, so that committing and searching further are both real options?}
{A candidate answer is stated or implied, and stopping at this point would produce an answer rather than a refusal.}
{No candidate has formed, the step is the first look at the image, or an answer was already committed earlier.} \\
\arrayrulecolor{black!15}\midrule\arrayrulecolor{black}
\rubitem{TrLook}{Thought--Action Faithfulness}{step, three-point ordinal}
{Does the reasoning at this step describe what is visible in the returned crop, and does it justify the action that follows?}
{\emph{Faithful} names objects, text or spatial relations present in the crop and ties them to the next action. \emph{Partial} does only one of the two.}
{\emph{Unfaithful} describes content absent from the crop, falls back on generic wording such as \textit{``examining the region''}, or contradicts the crop.} \\
\arrayrulecolor{black!15}\midrule\arrayrulecolor{black}
\rubitem{TrRex}{Pattern Fidelity}{re-examination, five-point Likert}
{How clearly does this re-examination follow its recorded pattern, one of attribute re-read, constraint conflict, missed-region sweep, reject re-verify, count reconcile or context expansion?}
{\emph{5} the recorded pattern is unambiguous, \emph{4} it is clear apart from one minor deviation, \emph{3} it is partially recognizable.}
{\emph{2} the pattern is weak or inconsistent, \emph{1} the re-examination does what a different pattern describes.} \\
\arrayrulecolor{black!15}\midrule\arrayrulecolor{black}
\rubitem{TrOk}{Adjudication Agreement}{decision, binary}
{Here are the search state and its two continuations, one that re-examines and one that commits. Which of them reaches the reference answer?}
{The choice made without seeing the label matches the adjudicated label of that decision.}
{Both continuations reach the same outcome, or the choice contradicts the adjudicated label.} \\
\bottomrule
\end{tabular}
\end{table}

\begin{table}[t]
\centering
\caption{Human evaluation of \resee{}. Favorable rate pools the $3n$ judgments, counting scores of 4 or 5 for pattern fidelity and agreement with the counterfactual label for adjudication. The final column reports agreement among raters.}
\label{tab:human}
\small
\setlength{\tabcolsep}{7pt}
\begin{tabular}{l l r c c}
\toprule
Dimension & Unit & $n$ & Favorable rate & Agreement \\
\midrule
Question validity & Trajectory & 120 & 93.3\% & $\kappa=0.73$ \\
Decision-state validity & Decision & 120 & 89.2\% & $\kappa=0.65$ \\
Thought--action faithfulness & Step & 240 & 76.4\% & $\alpha=0.61$ \\
Pattern fidelity & Re-examination & 120 & 73.6\% & $\alpha=0.79$ \\
Adjudication agreement & Decision & 120 & 83.3\% & $\kappa=0.73$ \\
\bottomrule
\end{tabular}
\end{table}

\subsection{Examples}
\label{app:showcase}
\begin{figure}[h]
\centering
\includegraphics[width=\textwidth]{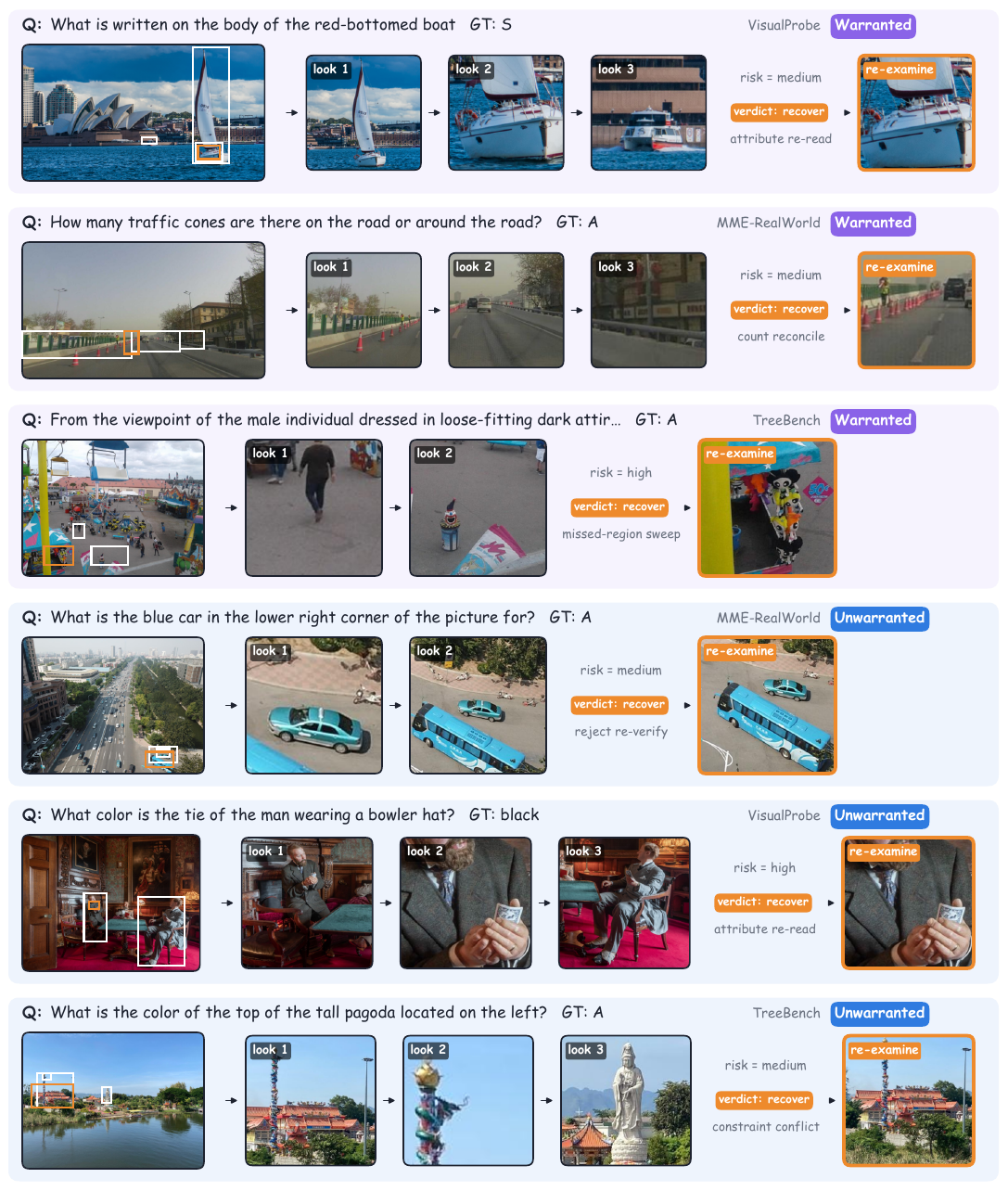}
\caption{Six \resee{} trajectories. Re-examination is warranted in the first three cases and unwarranted in the last three.}
\label{fig:showcase}
\end{figure}

Figure~\ref{fig:showcase} shows six trajectories of \resee{}, each with a single re-examination. Boxes mark the inspected regions, the re-examined one in orange, and the observations appear in the order requested. The annotator proposes re-examination in all six cases and rates its risk as medium or high, yet only the first three change the outcome.

\section{Details of \care{}}
\label{app:care}

\subsection{Training Procedure}
\label{app:algorithm}
Algorithm~\ref{alg:care} summarizes one training iteration of \care{} in the notation of Section~\ref{sec:method}.

\paragraph{Locating the Verdict Span.}\label{app:spans} The verdict is an attribute of the \tagname{lookback} tag, so its text is found by matching that attribute in the decoded response. The matched characters are then mapped to token positions on the sequence over which the loss is computed, that is, after the placeholders of the returned observations have been expanded. Each decision yields exactly one span, which contains neither the surrounding rationale nor the answer.

\subsection{Cold-Start Supervision}
\label{app:corpora}
All six variants use the same $2{,}197$ episodes and change re-examination supervision. \emph{Full} retains all $450$ demonstrated re-examinations across $437$ episodes. \emph{Pruned} removes those adjudicated as unwarranted and changes the deciding verdict to \textsc{skip}, leaving $95$; both variants retain the same $5{,}883$ ordinary looks. \emph{Masked} retains the full text but excludes unwarranted re-examinations from the loss. \emph{Stripped} removes the protocol tags while retaining the reasoning. \emph{Single} uses the one-turn form. \emph{External} is a size-matched external cold start re-expressed in our tag vocabulary; its trajectories are $3.4\times$ shallower and contain no re-examination decisions.

\subsection{Decision-Level Alignment}
\label{app:alignment-proof}
\noindent\textbf{Proof of Proposition~\ref{prop:decision-alignment}.}
Fix $h$ and write $q_v=Q^\pi(h,v)$ for the mean continuation reward and $\sigma_v^2$ for its finite variance under verdict $v$. The continuation policies are held fixed while differentiating the current verdict probability. Write $c=c(h)$ and $\tau=q_{\mathrm{rec}}-q_{\mathrm{skip}}-\lambda c$. Let $R_v$ be the on-policy return conditional on the sampled verdict $v$, and let $R_{\neg v}^{(1)},\ldots,R_{\neg v}^{(k)}$ be independent returns from resets to the same state with the opposite verdict imposed. We assume that a forced replay has the same conditional return law as an on-policy continuation given that verdict. This requires the same model prefix and decoding settings after the forced verdict, with fresh sampling randomness. The raw paired contrast is
\begin{equation*}
D_v=\begin{cases}
R_{\mathrm{rec}}-k^{-1}\sum_{j=1}^{k}R_{\mathrm{skip}}^{(j)}-\lambda c,&v=\textsc{recover},\\
k^{-1}\sum_{j=1}^{k}R_{\mathrm{rec}}^{(j)}-R_{\mathrm{skip}}-\lambda c,&v=\textsc{skip}.
\end{cases}
\end{equation*}
Exact reset and the conditional replay law place both branches at $h$ under the corresponding verdict distributions, so $\mathbb E[D_v\mid h,v]=\tau$ for either verdict. Conditional independence makes the covariance between the observed and replayed returns zero and makes the variance of the replay average $\sigma_{\neg v}^2/k$. Hence $\operatorname{Var}(D_v\mid h,v)=\sigma_v^2+\sigma_{\neg v}^2/k$.

For the binary verdict abstraction, $\partial_z\log\pi(\textsc{recover}\mid h)=1-p$ and $\partial_z\log\pi(\textsc{skip}\mid h)=-p$. The \care{} verdict credit is $\beta s(v)D_v$, with $s(\textsc{recover})=1$ and $s(\textsc{skip})=-1$. Taking expectation over the sampled verdict and then over its two continuations gives
\begin{align*}
\mathbb E\!\left[\beta s(v)D_v\,\partial_z\log\pi(v\mid h)\mid h\right]
&=\beta p(1-p)\mathbb E[D_{\mathrm{rec}}\mid h,\textsc{recover}]\\
&\quad+\beta(1-p)p\mathbb E[D_{\mathrm{skip}}\mid h,\textsc{skip}]\\
&=2\beta p(1-p)\tau.
\end{align*}
The cost-adjusted decision value, with continuation policies fixed, is $J_h(z)=p(z)(q_{\mathrm{rec}}-\lambda c)+(1-p(z))q_{\mathrm{skip}}$. Since $\partial_z p=p(1-p)$, its derivative is $\partial_zJ_h=p(1-p)\tau$. The raw \care{} verdict term therefore has the same sign as this local derivative for $p\in(0,1)$ and $\beta>0$.

For the implemented clipping, let $C_v=\mathrm{clip}(D_v,-\kappa,\kappa)$ and define
\begin{equation*}
B_\kappa(h)=\frac{1}{2}\sum_{v\in\{\textsc{recover},\textsc{skip}\}}
\mathbb E\!\left[(|D_v|-\kappa)_+\mid h,v\right].
\end{equation*}
For every real $d$, $|\mathrm{clip}(d,-\kappa,\kappa)-d|=(|d|-\kappa)_+$. Write $e_v=\mathbb E[C_v-D_v\mid h,v]$. Replacing $D_v$ by $C_v$ in the score identity gives
\begin{equation*}
\mathbb E\!\left[\beta s(v)C_v\,\partial_z\log\pi(v\mid h)\mid h\right]
=2\beta p(1-p)\left(\tau+\frac{e_{\mathrm{rec}}+e_{\mathrm{skip}}}{2}\right).
\end{equation*}
The triangle inequality and $|\mathbb E X|\le\mathbb E|X|$ give
\begin{equation*}
\left|\frac{e_{\mathrm{rec}}+e_{\mathrm{skip}}}{2}\right|
\le\frac12\sum_v\mathbb E\!\left[|C_v-D_v|\mid h,v\right]
=B_\kappa(h).
\end{equation*}
The finite second moments assumed in the proposition ensure that these expectations exist. For $|\tau|>B_\kappa(h)$, the perturbation cannot change the sign of $\tau$, proving the clipping claim. A complete verdict span has this binary score if its two allowed sequences exhaust the normalized choice under a fixed prefix.\hfill$\square$

\begin{corollary}[Finite-Replay Direction Reliability]\label{cor:replay-reliability}
Under the assumptions of Proposition~\ref{prop:decision-alignment}, let $\tau=\tau^\pi(h)\ne0$, $\beta,\kappa>0$, and $V_v(k)=\sigma_v^2+\sigma_{\neg v}^2/k$. For the clipped verdict score $G_v=\beta s(v)\mathrm{clip}(D_v,-\kappa,\kappa)\,\partial_z\log\pi(v\mid h)$,
\begin{equation}
\Pr\!\left(\tau G_v\le0\mid h,v\right)
\le\frac{V_v(k)}{V_v(k)+\tau^2}.
\label{eq:replay-reliability}
\end{equation}
The upper bound is nonincreasing in the replay count and converges to $\sigma_v^2/(\sigma_v^2+\tau^2)$.
\end{corollary}

\noindent\textbf{Proof.}
Conditional on $v$, the multiplier $s(v)\partial_z\log\pi(v\mid h)$ equals $1-p$ for \textsc{recover} and $p$ for \textsc{skip}. Both are strictly positive. Symmetric clipping with $\kappa>0$ preserves the sign of each realized contrast, so $\tau G_v\le0$ is equivalent to $\tau D_v\le0$. Set $t=|\tau|$ and $X=-\operatorname{sign}(\tau)(D_v-\tau)$. Proposition~\ref{prop:decision-alignment} gives $\mathbb E[X\mid h,v]=0$ and $\mathbb E[X^2\mid h,v]=V_v(k)$, and the adverse event is $X\ge t$.

For every $a\ge0$, the event $X\ge t$ implies $(X+a)^2\ge(t+a)^2$. Applying Markov's inequality to the nonnegative random variable $(X+a)^2$ yields
\begin{equation*}
\Pr(X\ge t\mid h,v)
\le\frac{\mathbb E[(X+a)^2\mid h,v]}{(t+a)^2}
=\frac{V_v(k)+a^2}{(t+a)^2}.
\end{equation*}
For $V_v(k)>0$, choose $a=V_v(k)/t$ and simplify the ratio to $V_v(k)/(V_v(k)+t^2)$. For $V_v(k)=0$, $X=0$ almost surely and $t>0$, so the event has probability zero. This proves Eq.~(\ref{eq:replay-reliability}). The function $x/(x+\tau^2)$ is increasing for $x\ge0$, and $V_v(k)$ is nonincreasing and converges to $\sigma_v^2$ as $k$ grows. Substitution gives the stated limit.\hfill$\square$

\begin{corollary}[Value Signal under Same-Verdict Replay]\label{cor:same-verdict}
Keep the state, binary score and continuation assumptions of Proposition~\ref{prop:decision-alignment}. Replace opposite-verdict replay by $k$ independent continuations under the observed verdict and define $D_v^{\mathrm{same}}=s(v)(R_v-k^{-1}\sum_j R_v^{(j)})-\lambda c(h)$. Then
\begin{equation}
\mathbb E\!\left[\beta s(v)D_v^{\mathrm{same}}\,\partial_z\log\pi(v\mid h)\mid h\right]
=-2\beta p(1-p)\lambda c(h).
\label{eq:same-verdict}
\end{equation}
For $k=1$, $\kappa>0$ and $\lambda c(h)\ge0$, the expected clipped score is nonpositive and equals zero at zero cost.
\end{corollary}

\noindent\textbf{Proof.}
All $k+1$ returns have conditional mean $q_v$, so
\begin{equation*}
\mathbb E[D_v^{\mathrm{same}}\mid h,v]
=s(v)\left(q_v-\frac1k\sum_{j=1}^k q_v\right)-\lambda c(h)
=-\lambda c(h).
\end{equation*}
Substitution into the two terms of the binary score expectation gives Eq.~(\ref{eq:same-verdict}). The difference $q_{\mathrm{rec}}-q_{\mathrm{skip}}$ has canceled because each return is compared with another continuation under the same action.

For $k=1$, write $Z_v=R_v-R_v^{(1)}$ and $a=\lambda c(h)$. Exchanging the independent, identically distributed returns changes $Z_v$ to $-Z_v$ without changing its law. Consequently, the odd function $\mathrm{clip}(\cdot,-\kappa,\kappa)$ has zero expectation under $Z_v$. Define $F_v(b)=\mathbb E[\mathrm{clip}(Z_v+b,-\kappa,\kappa)\mid h,v]$. This expectation is finite since clipping is bounded. Pointwise monotonicity gives $F_v(-a)\le F_v(0)=0\le F_v(a)$ for $a\ge0$. Oddness also gives
\begin{equation*}
s(v)\mathrm{clip}(s(v)Z_v-a,-\kappa,\kappa)
=\mathrm{clip}(Z_v-a\,s(v),-\kappa,\kappa).
\end{equation*}
The expected clipped score is therefore $\beta p(1-p)[F_{\mathrm{rec}}(-a)-F_{\mathrm{skip}}(a)]\le0$. At $a=0$, both terms vanish.\hfill$\square$

Opposite-verdict pairing retains the difference in decision values that same-verdict replay cancels.

\begin{algorithm}[!b]
\caption{One training iteration of \care{}}
\label{alg:care}
\small
\begin{algorithmic}[1]
\Require policy $\pi_\theta$ initialized by the cold start on \resee{}, prompts $\mathcal D$, group size $G$, replays $k$, maximum decisions $M$, price $\lambda$, bound $\kappa$, weights $\beta, \gamma, \eta$
\Ensure updated policy $\pi_\theta$
\For{each prompt $x \in \mathcal D$}
  \algphase{Rollout}
  \State sample a group of rollouts $\{\zeta_i\}_{i=1}^{G} \sim \pi_{\mathrm{old}}(\cdot \mid x)$ and score $R_i = R(\zeta_i)$
  \algphase{Paired intervention}
  \For{each rollout $\zeta_i$ and each of its first $M$ decision states $h_{i,j}$}
    \State read the written verdict $v_{i,j}$ and its token positions $\mathcal V_{i,j}$ \Comment{located from the protocol}
    \State reset the environment to $h_{i,j}$ and impose the opposite verdict $\neg v_{i,j}$ \Comment{exact reset}
    \State sample $k$ replays from $\pi_{\mathrm{old}}$ and average their rewards into $\bar R^{\,\mathrm{cf}}_{i,j}$ \Comment{no gradient}
    \State set $(\bar R_{\mathrm{rec}}, \bar R_{\mathrm{skip}}) \gets (R_i, \bar R^{\,\mathrm{cf}}_{i,j})$ if $v_{i,j} = \textsc{recover}$, and $(\bar R^{\,\mathrm{cf}}_{i,j}, R_i)$ otherwise
    \State \colorbox{ReseeOrange!14}{$\hat\tau_{i,j}$} $\gets \mathrm{clip}\big(\bar R_{\mathrm{rec}} - \bar R_{\mathrm{skip}} - \lambda c(h_{i,j}),\, -\kappa,\, \kappa\big)$ \Comment{Eq.~(\ref{eq:tau-hat})}
  \EndFor
  \algphase{Localized credit}
  \For{each token position $t$ of each rollout $\zeta_i$}
    \State \colorbox{ReseeBlue!12}{$\tilde A_{i,t}$} $\gets \beta\, s(v_{i,j})\, \hat\tau_{i,j}$ if $t \in \mathcal V_{i,j}$, and $(R_i - \mu_{g(i)}) / (\sigma_{g(i)} + \delta)$ otherwise \Comment{Eq.~(\ref{eq:adv})}
  \EndFor
\EndFor
\algphase{Policy update}
\State update $\theta$ by descending $\mathcal L(\theta)$ on the rollouts $\{\zeta_i\}$ only \Comment{Eq.~(\ref{eq:loss})}
\end{algorithmic}
\end{algorithm}

\section{Experimental Setup}
\label{app:setup}

\subsection{Evaluation Protocol}
\label{app:eval}
\paragraph{Answer Grading.} Answers are graded by exact normalized match, followed by a bracketed-letter match when the reference is a single letter. Free-text references also allow word-bounded containment. An optional LLM stage sees the question and response, without the reference, and may rescue answers marked incorrect by the rules. Both scores are reported. The two untrained baselines may omit the \tagname{answer} tag, while trained policies must emit it. $\Delta$Acc is reported when both sides contain at least twenty records. Table~\ref{tab:main} and the framework analyses use our $980$-question split, six-source macro average, and a common six-round generation budget. Table~\ref{tab:bench} uses the arithmetic mean of five displayed sources. VisualProbe is excluded from that comparison because its free-text references do not match the option template.

\paragraph{Macro Aggregation.} Let $a_{b,r}$ be the accuracy on source $b$ in evaluation run $r$. The six-source macro accuracy, its mean over 32 runs, and its sample standard deviation are
\begin{equation*}
\begin{aligned}
A_r &= \frac{1}{6}\sum_{b=1}^{6}a_{b,r}, \qquad
\bar A = \frac{1}{32}\sum_{r=1}^{32}A_r,\\
s_A &= \left[\frac{1}{31}\sum_{r=1}^{32}(A_r-\bar A)^2\right]^{1/2}.
\end{aligned}
\end{equation*}
Table~\ref{tab:main} reports $\bar A$ and $s_A$. Aggregating sources within each run preserves their covariance in the reported spread.

\paragraph{Decision Metrics.}\label{app:metrics} We score the $288$ adjudicated evaluation decisions against their labels. A re-examination counts as taken when the policy executes a \tagname{recover} action at that state, since \tagname{lookback} is emitted at every decision, including commitments. Writing $D$ for these decisions, $W \subseteq D$ for the warranted ones and $\hat W \subseteq D$ for those at which the policy re-examines,
\begin{equation*}
\mathrm{recall} = \frac{|\hat W \cap W|}{|W|},
\qquad
\mathrm{precision} = \frac{|\hat W \cap W|}{|\hat W|},
\qquad
\mathrm{over\text{-}examination} = \frac{|\hat W \setminus W|}{|D \setminus W|},
\end{equation*}
and balanced accuracy is the mean of recall and $1 - \mathrm{over\text{-}examination}$. Always re-examining gives $100\%$ recall, $18.1\%$ precision and $100\%$ over-examination. Never re-examining gives zero recall and over-examination, with undefined precision. Both policies have $50\%$ balanced accuracy. We pool decision metrics across sources, which contain between one and sixteen positive cases each. Grounding IoU is reported separately and its reward uses independently annotated boxes.

\subsection{Unseen Benchmarks}
\label{app:ood-benchmarks}
\paragraph{High-Resolution Perception.} HR-Bench 8K~\citep{wang2025divide} contains 200 questions on fine-grained perception in 8K images, split equally between single-instance properties and cross-instance relations. MagnifierBench~\citep{li2023otterhd} contains 283 questions over 166 images and tests small-object identification, counting and attributes that often require a closer view. XLRS-Bench-Lite~\citep{wang2025xlrs} contributes 2,810 multiple-choice VQA items from ultra-high-resolution remote-sensing imagery, extending the evaluation to land-use, object and spatial reasoning.

\paragraph{Extended Visual Search.} We use the initial 350-task release of VistaHop~\citep{he2026vistahop}, which tests repeated inspection and multi-hop reasoning over visual anchors. VisualNeedle~\citep{chen2026visualneedle} tests crop-based search for small clues in 300 questions covering text, attributes, relations and occlusion. Visual Haystacks~\citep{wu2025visual} tests retrieval across unrelated images. We evaluate its 1,000 binary questions in the 50-image single-needle setting, with one anchor-bearing image among distractors.

\paragraph{Scoring and Observations.} VistaHop uses VistaArena Pass@1 and the other benchmarks report accuracy. VisualNeedle follows its tool-enabled protocol. We report request counts by benchmark from the harness traces. All six evaluation sets are disjoint from \resee{} training.

\subsection{Compared Methods}
\label{app:baselines}
\paragraph{Trained Policies.} Within each backbone, the trained comparisons in Table~\ref{tab:main} share the RL data and reward. All use the \emph{Full} cold start except GRPO-Pruned, which uses the \emph{Pruned} variant. The methods differ as follows.
\begin{itemize}[leftmargin=14pt,itemsep=1pt,topsep=3pt]
\item \textbf{Always-prompt, Never-prompt, Random.} The cold-started backbone instructed to favor \textsc{recover}, favor \textsc{skip}, or select \textsc{recover} with a fixed random probability. We score the actions each policy executes.
\item \textbf{Zero-shot, Prompted.} The untrained backbone under the common contract and under the selective prompt of Appendix~\ref{app:bench}.
\item \textbf{SFT.} The cold-start policy without reinforcement learning.
\item \textbf{GRPO}~\citep{shao2024deepseekmathpushinglimitsmathematical}. Group-relative policy optimization, which normalizes the trajectory reward within a group of rollouts and assigns the resulting advantage to every token.
\item \textbf{GiGPO}~\citep{feng2026group}. A two-level advantage that adds to the episode-level group advantage a step-level one, obtained by grouping actions taken from the same environment state across rollouts.
\item \textbf{Entropy}~\citep{wang2026beyond}. GRPO with policy-gradient updates restricted to the $20\%$ of tokens with the highest entropy.
\item \textbf{GRPO-Pruned.} GRPO from the \emph{Pruned} cold start, which removes the demonstrated re-examinations adjudicated as unwarranted and therefore uses the labels of \resee{}.
\end{itemize}

\paragraph{Ablation Protocol.} Table~\ref{tab:ablation} uses Qwen3-VL-8B. The verdict-flip ablation replays the observed verdict while retaining the branch ordering and cost term of Eq.~(\ref{eq:tau-hat}), with the same $k=1$ replay and clipping as the main setting. The localization ablation assigns the signed contrast throughout the trajectory, while the trajectory-cost variant moves the observation penalty from the paired contrast into the trajectory reward.

\paragraph{Released Agents.} We evaluate the agents in Table~\ref{tab:bench} from their public checkpoints and include their shared backbone.
\begin{itemize}[leftmargin=14pt,itemsep=1pt,topsep=3pt]
\item \textbf{Pixel Reasoner}~\citep{su2026pixel}. Instruction tuning on synthesized traces with zoom-in and select-frame operations, followed by RL with a curiosity-driven reward.
\item \textbf{DeepEyes}~\citep{zheng2026deepeyes}. End-to-end RL without a cold start, in which the model crops regions with its own grounding ability.
\item \textbf{ViGoRL}~\citep{sarch2026grounded}. RL that anchors each reasoning step to image coordinates, with a multi-turn variant that zooms into the predicted coordinates.
\item \textbf{TreeVGR}~\citep{wang2026traceable}. RL that supervises localization and reasoning jointly through a box-level reward.
\item \textbf{Mini-o3}~\citep{lai2026mini}. A diverse cold start followed by RL with over-turn masking, which scales visual search to tens of interaction turns.
\item \textbf{Qwen2.5-VL-7B}~\citep{bai2025qwen25vltechnicalreport}. The instruction-tuned backbone of the five agents above, without any visual-search training.
\end{itemize}

\section{Existing VLMs on \resee{}}
\label{app:vlms}

\subsection{Models, Prompts and Full Results}
\label{app:bench}
\paragraph{Query Setup.} We query the eight models of Section~\ref{sec:bench-results} through the OpenRouter API in their instruction-tuned public releases, with at most $1{,}024$ output tokens per turn, the provider's default temperature and twelve rounds. Each turn returns the requested crop, and a trajectory ends at the first answer. We exclude trajectories abandoned by the API after repeated time-outs from scoring. Table~\ref{tab:bench-full} reports every model under every prompt.

\paragraph{Prompts.} All prompts except the selective one extend a common contract that defines \tagname{look}, \tagname{recover} and \tagname{answer} with boxes normalized to $[0,1]$ and requires reasoning inside \tagname{think}. The default prompt asks the model to reason about the evidence already seen, the requirements of each remaining option and the constraint the next action should settle. The always and never prompts require at least one \tagname{recover} or forbid it. The random prompt tells the model, per question and with probability $0.22$ matched to the annotated re-examination rate, whether it must re-examine or decide from the crops already seen. The selective prompt uses a plain protocol description and asks the model to re-examine when the evidence does not settle the question, noting that looking back costs a turn.

\begin{promptbox}{Common Contract, Shared by All Prompts Except the Selective One}
You are performing visual search on an image to answer a multiple-choice question. You examine the image by requesting crops; the environment returns the requested crop each round. You may also re-examine a region you have already seen.

You may reason before each action, but put all reasoning inside \tthink{<think>} tags. Never mention any tag name in your reasoning text.

Format -- one action per turn:
\begin{itemize}[leftmargin=12pt,itemsep=2pt,topsep=3pt,label={\color{ReseeBlue!60}\scriptsize$\blacktriangleright$}]
\item \tlook{<look bbox="[x0, y0, x1, y1]"/>}\enspace Request a crop. Coordinates are normalized [0,1]: 0 = left/top, 1 = right/bottom, with x0\,$<$\,x1 and y0\,$<$\,y1.
\item \trecover{<recover pattern="..." bbox="[x0, y0, x1, y1]"/>}\enspace Go back and re-examine a region. pattern is a short reason for the recheck.
\item \tanswer{<answer>X</answer>}\enspace Give the final answer, where X is the chosen option letter.
\end{itemize}
Continue issuing look/recover actions until you can answer, then output \tanswer{<answer>}.
\end{promptbox}

\begin{promptbox}{Default Prompt: Explicit Reasoning, Appended to the Contract}
Before each action, work through the evidence explicitly inside \tthink{<think>}: what you have already seen, what each remaining option would require, and which unresolved constraint your next action is meant to settle. Then emit the action.
\end{promptbox}

\begin{promptbox}{Selective Prompt: Protocol Description That Replaces the Contract}
You answer questions about a high-resolution image by searching it.

Think inside \tthink{<think>} and \tthink{</think>}. To inspect a region more closely, emit \tlook{<look bbox="[x0, y0, x1, y1]"/>} and stop; a crop of that region is returned to you as the next image. Coordinates are normalized to the image, between 0 and 1.

\begin{tcolorbox}[enhanced,arc=1mm,boxrule=0pt,colback=ReseeOrange!8,left=5pt,right=5pt,top=3pt,bottom=3pt,before skip=4pt,after skip=4pt,fontupper=\footnotesize]
Before committing, judge whether the evidence you have already gathered settles the question. Go back and re-examine a region with \trecover{<recover pattern="..." bbox="[x0, y0, x1, y1]"/>} when it does not. Looking back costs a turn, so do it only when it would change your answer.
\end{tcolorbox}
Put the final answer inside \tanswer{<answer>} and \tanswer{</answer>}.
\end{promptbox}

\begin{promptbox}{Control Prompts: Instructions Appended to the Contract}
\begin{tcbraster}[raster columns=2,raster equal height,raster column skip=6pt,raster before skip=2pt,raster after skip=4pt]
\begin{minicard}{ReseeBlue}{Always Re-examine}
After your first look, always go back and re-examine a region you have already seen before you answer. Take at least one \trecover{<recover>} action on every question, even when you believe you already have enough evidence.
\end{minicard}
\begin{minicard}{ReseeOrange}{Never Re-examine}
Never go back to a region you have already examined. Do not emit \trecover{<recover>}. Decide from the crops you have already been shown and answer as soon as you can.
\end{minicard}
\end{tcbraster}
\begin{minicard}{ReseeSlate}{Random Verdict, Drawn Once per Question}
A coin has been flipped for this question. You will be told below whether you are in a re-examine question or a decide-now question; follow that instruction.\par\smallskip
{\color{ReseeBlue!85!black}\scriptsize\bfseries With probability 0.22.}\enspace This is a re-examine question: take a \trecover{<recover>} action on a region you have already seen before you answer.\par\smallskip
{\color{ReseeOrange!85!black}\scriptsize\bfseries With probability 0.78.}\enspace This is a decide-now question: do not \trecover{<recover>}; answer from the crops you have already been shown.
\end{minicard}
\end{promptbox}

\begin{figure}[h]
\centering
\includegraphics[width=\textwidth]{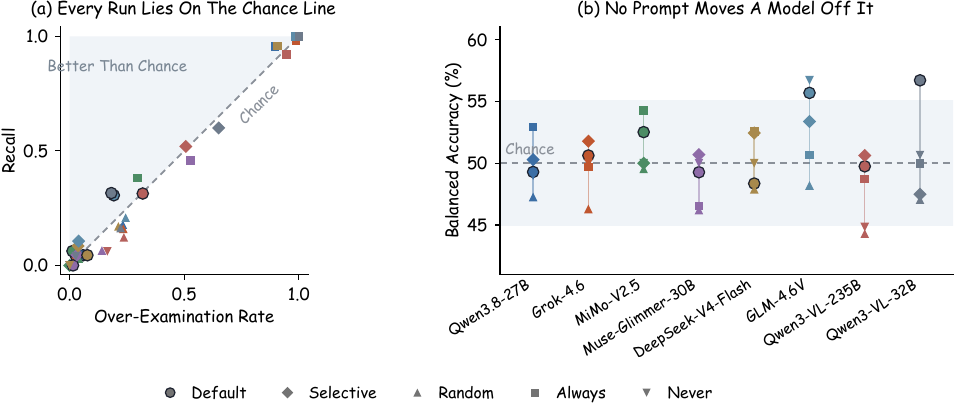}
\caption{Eight VLMs under five prompts on 288 adjudicated decisions. Panel (a) compares recall with over-examination, with default prompts outlined. Panel (b) shows balanced accuracy against the $50\%$ chance level.}
\label{fig:bench}
\end{figure}

\begin{table}[h]
\centering
\caption{All model and prompt combinations on \resee{}. Of the $n$ scorable adjudicated decisions, ``Pos.'' are warranted. ``Failed'' counts incomplete provider trajectories.}
\label{tab:bench-full}
\resizebox{\textwidth}{!}{%
\begin{tabular}{l c ccccc c rrr}
\toprule
 & \textbf{Answer} & \multicolumn{5}{c}{\textbf{Re-examination decision}} & \textbf{Behaviour} & \multicolumn{3}{c}{\textbf{Coverage}} \\
\cmidrule(lr){2-2}\cmidrule(lr){3-7}\cmidrule(lr){8-8}\cmidrule(lr){9-11}
\textbf{Prompt} & Acc.\ $\uparrow$ & Recall $\uparrow$ & Prec.\ $\uparrow$ & Over-ex.\ $\downarrow$ & F1 $\uparrow$ & Bal.\ acc.\ $\uparrow$ & Re-ex. & $n$ & Pos. & Failed \\
\midrule
\rowcolor{ReseeBand}\multicolumn{11}{l}{\textbf{Qwen3.8-27B}} \\
Default &\underline{73.1}&4.4&14.3&5.9&6.8&\cellcolor{ReseeRed!10}49.3& 0.06 & 250 & 45 & 54 \\
Selective &\textbf{75.3}&4.7&\textbf{20.0}&\underline{4.1}&7.5&\cellcolor{ReseeBlue!9}\underline{50.3}& 0.03 & 239 & 43 & 76 \\
Random &70.8&\underline{17.8}&14.8&23.2&\underline{16.2}&\cellcolor{ReseeRed!18}47.3& 0.27 & 243 & 45 & 66 \\
Always &72.6&\textbf{95.6}&\underline{18.4}&89.7&\textbf{30.8}&\cellcolor{ReseeBlue!19}\textbf{52.9}& 1.03 & 258 & 45 & 43 \\
Never &71.3&0.0&0.0&\textbf{2.0}& -- &\cellcolor{ReseeRed!12}49.0& 0.02 & 243 & 43 & 65 \\
\rowcolor{ReseeBand}\multicolumn{11}{l}{\textbf{Grok-4.6}} \\
Default &70.2&4.0&\underline{25.0}&2.8&6.9&\cellcolor{ReseeBlue!10}\underline{50.6}& 0.02 & 268 & 50 & 28 \\
Selective &69.0&5.8&\textbf{37.5}&\underline{2.2}&10.0&\cellcolor{ReseeBlue!15}\textbf{51.8}& 0.03 & 278 & 52 & 14 \\
Random &69.1&\underline{16.0}&13.6&23.4&\underline{14.7}&\cellcolor{ReseeRed!22}46.3& 0.23 & 268 & 50 & 24 \\
Always &\textbf{71.5}&\textbf{98.1}&18.3&98.7&\textbf{30.9}&\cellcolor{ReseeRed!9}49.7& 1.01 & 282 & 52 & 12 \\
Never &\underline{71.4}&0.0& -- &\textbf{0.0}& -- &\cellcolor{ReseeBlue!8}50.0& 0.00 & 267 & 49 & 28 \\
\rowcolor{ReseeBand}\multicolumn{11}{l}{\textbf{MiMo-V2.5}} \\
Default &\underline{69.3}&\underline{6.2}&\textbf{33.3}&\underline{1.2}&\underline{10.5}&\cellcolor{ReseeBlue!18}\underline{52.5}& 0.01 & 180 & 16 & 191 \\
Selective &\textbf{69.9}&0.0& -- &\textbf{0.0}& -- &\cellcolor{ReseeBlue!8}50.0& 0.00 & 206 & 28 & 157 \\
Random &66.3&3.0&12.5&4.0&4.9&\cellcolor{ReseeRed!9}49.5& 0.05 & 210 & 33 & 126 \\
Always &67.6&\textbf{38.2}&\underline{21.0}&29.7&\textbf{27.1}&\cellcolor{ReseeBlue!25}\textbf{54.3}& 0.24 & 199 & 34 & 158 \\
Never &69.0&0.0& -- &\textbf{0.0}& -- &\cellcolor{ReseeBlue!8}50.0& 0.00 & 184 & 28 & 163 \\
\rowcolor{ReseeBand}\multicolumn{11}{l}{\textbf{Muse-Glimmer-30B}} \\
Default &66.0&0.0&0.0&\underline{1.5}& -- &\cellcolor{ReseeRed!10}49.3& 0.02 & 251 & 45 & 54 \\
Selective &\textbf{69.9}&4.5&\textbf{25.0}&3.2&\underline{7.7}&\cellcolor{ReseeBlue!10}\textbf{50.7}& 0.03 & 234 & 44 & 99 \\
Random &\underline{66.7}&\underline{6.5}&9.4&14.1&\underline{7.7}&\cellcolor{ReseeRed!23}46.2& 0.36 & 252 & 46 & 63 \\
Always &63.6&\textbf{45.8}&\underline{15.9}&52.7&\textbf{23.7}&\cellcolor{ReseeRed!21}46.6& 1.12 & 268 & 48 & 36 \\
Never &66.4&0.0& -- &\textbf{0.0}& -- &\cellcolor{ReseeBlue!8}\underline{50.0}& 0.00 & 246 & 44 & 71 \\
\rowcolor{ReseeBand}\multicolumn{11}{l}{\textbf{DeepSeek-V4-Flash}} \\
Default &\underline{65.7}&4.4&11.1&7.7&6.3&\cellcolor{ReseeRed!14}48.4& 0.22 & 252 & 45 & 63 \\
Selective &\textbf{65.9}&8.5&\textbf{33.3}&\underline{3.6}&13.6&\cellcolor{ReseeBlue!17}\underline{52.4}& 0.11 & 267 & 47 & 28 \\
Random &61.7&\underline{17.0}&15.1&21.2&\underline{16.0}&\cellcolor{ReseeRed!16}47.9& 0.40 & 259 & 47 & 49 \\
Always &61.1&\textbf{95.8}&\underline{19.0}&90.7&\textbf{31.7}&\cellcolor{ReseeBlue!18}\textbf{52.5}& 2.56 & 264 & 48 & 42 \\
Never &60.7&0.0& -- &\textbf{0.0}& -- &\cellcolor{ReseeBlue!8}50.0& 0.00 & 249 & 45 & 51 \\
\rowcolor{ReseeBand}\multicolumn{11}{l}{\textbf{GLM-4.6V}} \\
Default &53.4&30.6&26.8&19.2&28.6&\cellcolor{ReseeBlue!30}\underline{55.7}& 0.21 & 262 & 49 & 37 \\
Selective &\underline{54.3}&10.6&\textbf{38.5}&\textbf{3.9}&16.7&\cellcolor{ReseeBlue!21}53.4& 0.05 & 252 & 47 & 54 \\
Random &47.4&20.8&15.6&24.4&17.9&\cellcolor{ReseeRed!15}48.2& 0.33 & 269 & 48 & 30 \\
Always &\textbf{55.5}&\textbf{100.0}&18.7&98.7&\textbf{31.5}&\cellcolor{ReseeBlue!10}50.7& 1.65 & 276 & 51 & 20 \\
Never &49.3&\underline{31.2}&\underline{28.3}&\underline{17.8}&\underline{29.7}&\cellcolor{ReseeBlue!34}\textbf{56.7}& 0.21 & 261 & 48 & 37 \\
\rowcolor{ReseeBand}\multicolumn{11}{l}{\textbf{Qwen3-VL-235B-A22B}} \\
Default &50.2&31.4&18.0&31.9&22.9&\cellcolor{ReseeRed!9}\underline{49.7}& 0.50 & 280 & 51 & 13 \\
Selective &\textbf{54.3}&\underline{51.9}&\textbf{19.3}&50.7&\underline{28.1}&\cellcolor{ReseeBlue!10}\textbf{50.6}& 1.04 & 275 & 52 & 17 \\
Random &52.7&12.2&10.3&\underline{23.6}&11.2&\cellcolor{ReseeRed!30}44.3& 0.39 & 269 & 49 & 29 \\
Always &\textbf{54.3}&\textbf{92.0}&\underline{18.1}&94.5&\textbf{30.3}&\cellcolor{ReseeRed!13}48.7& 1.99 & 270 & 50 & 25 \\
Never &\underline{53.0}&6.1&7.7&\textbf{16.4}&6.8&\cellcolor{ReseeRed!28}44.8& 0.27 & 268 & 49 & 30 \\
\rowcolor{ReseeBand}\multicolumn{11}{l}{\textbf{Qwen3-VL-32B}} \\
Default &44.7&31.6&\textbf{33.3}&\underline{18.2}&\textbf{32.4}&\cellcolor{ReseeBlue!34}\textbf{56.7}& 0.11 & 85 & 19 & 444 \\
Selective &\textbf{55.5}&\underline{60.0}&17.1&65.0&26.7&\cellcolor{ReseeRed!18}47.5& 1.83 & 273 & 50 & 21 \\
Random &45.7&16.3&13.8&22.2&15.0&\cellcolor{ReseeRed!19}47.1& 1.50 & 274 & 49 & 21 \\
Always &\underline{46.7}&\textbf{100.0}&18.5&100.0&\underline{31.2}&\cellcolor{ReseeBlue!8}50.0& 1.72 & 281 & 52 & 12 \\
Never &45.6&4.0&\underline{25.0}&\textbf{2.7}&6.9&\cellcolor{ReseeBlue!10}\underline{50.7}& 0.04 & 273 & 50 & 21 \\
\bottomrule
\end{tabular}}
\end{table}

\subsection{Decision Showcase}
\label{app:showcase-vlm}
\begin{figure}[h]
\centering
\includegraphics[width=\textwidth]{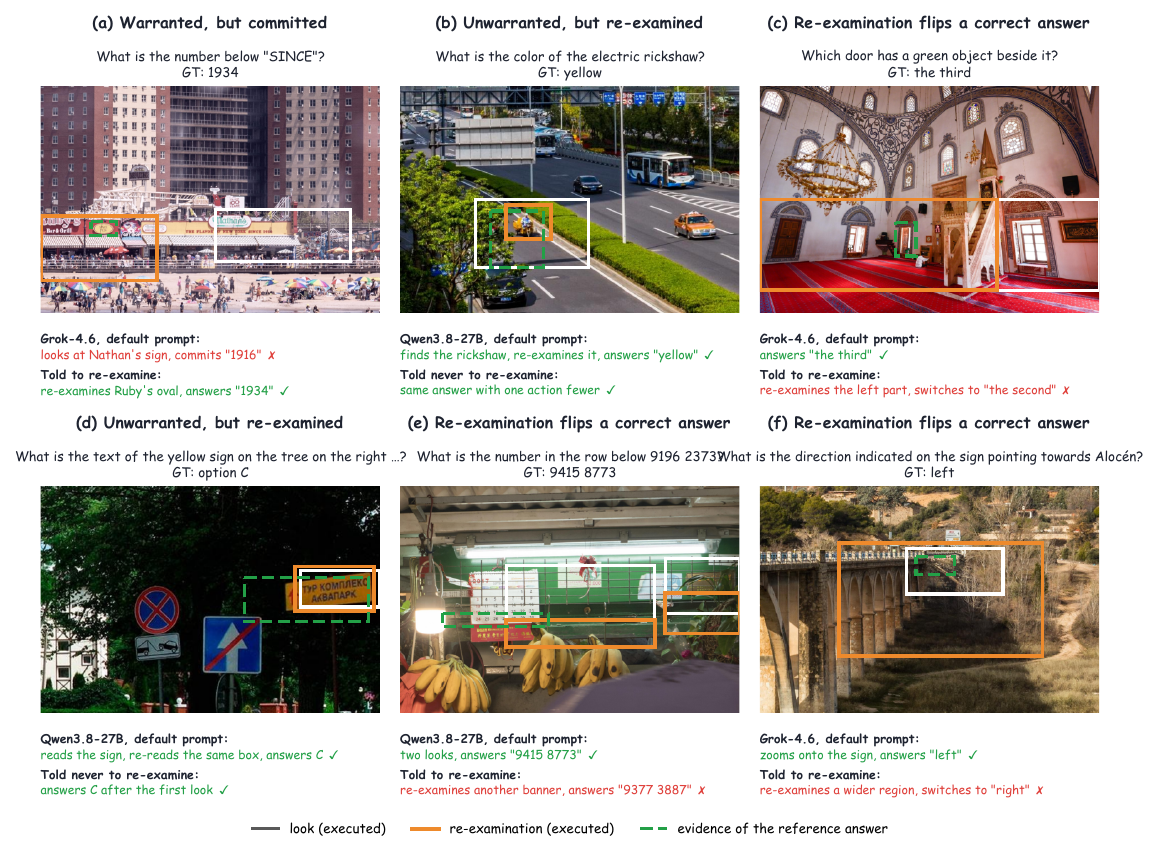}
\caption{Six decisions from existing VLMs. White and orange boxes mark executed looks and re-examinations, while dashed green boxes mark the reference evidence.}
\label{fig:cases}
\end{figure}

Figure~\ref{fig:cases} shows six decisions from Section~\ref{sec:bench-results} with actions quoted from the trajectories. A re-examination requires an executed \tagname{recover} action. Each model and prompt is a single sampled run, and the adjudication label refers to the decision in the reference trajectory.

In case (a), Grok-4.6 initially commits to \texttt{1916} after inspecting one of two signs marked ``SINCE''. Re-examination of the other sign yields \texttt{1934}. In cases (b) and (d), Qwen3.8-27B revisits a region after the answer is settled, adding an unnecessary action. In cases (c), (e) and (f), additional observations overturn correct answers by shifting attention to a different region. These cases illustrate both missed warranted re-examinations and unnecessary ones.

\section{Case Studies of the Trained Policy}
\label{app:traj}

\subsection{Full Trajectories for the Main-Text Cases}
\label{app:case-full}
\begin{figure}[p]
\centering
\begin{trajbox}{ReseeSuccess}{Route Planning: Resolving the Final Transfer}
\includegraphics[width=\linewidth]{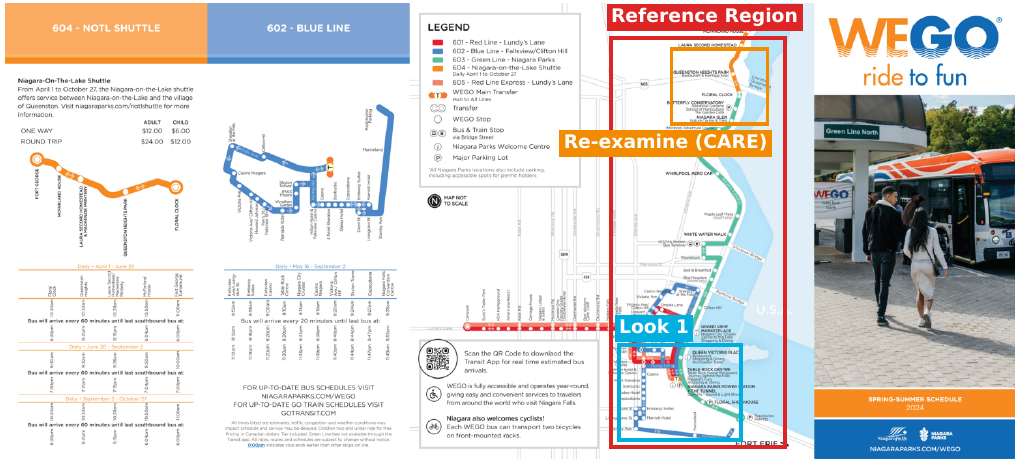}
\begin{trrow}{TrAsk}\trstep{TrAsk}{Question.} If I am staying at the Marriott Hotel and plan to go to Queenston Heights Park, can I reach it by transferring between buses without walking in between, and how many lines must I take at minimum? (A) No, 2. (B) No, 3. (C) Yes, 1. (D) Yes, 2. (E) Yes, 3. (F) None of the above.\end{trrow}
\begin{trrow}{TrPlan}\trstep{TrPlan}{Initial Scene.} The map separates the hotel district in the south from Queenston Heights Park in the north. A valid answer requires identifying the hotel's line, the destination's line and a connected sequence of transfer stops.\end{trrow}
\begin{trrow}{TrLook}\trstep{TrLook}{Shared Observation.} The southern crop places Marriott Hotel on the Blue Line (602). The Blue Line reaches Table Rock Centre, where it connects to the Green Line (603). This observation establishes the first transfer but leaves the destination connection unresolved.\end{trrow}
\begin{trrow}{TrOk}\trstep{TrOk}{CARE Verdict.} Re-examine the northern map before answering. The outstanding relation is whether the Green Line reaches the park or requires another transfer.\end{trrow}
\begin{trrow}{TrRex}\trstep{TrRex}{CARE Observation.} The Green Line reaches Floral Clock. A transfer there connects to the Niagara-on-the-Lake Shuttle (604), which serves Queenston Heights Park. The two transfer locations connect Blue, Green and Shuttle without a walking segment.\end{trrow}
\begin{trrow}{TrOk}\trstep{TrOk}{CARE Answer.} Commit to E, three lines. The additional observation supplies the missing transfer and completes the route.\end{trrow}
\begin{trrow}{TrBad}\trstep{TrBad}{GRPO Verdict and Answer.} Commit from the shared southern observation and answer D, two lines. This continuation treats the Green Line as serving the destination and omits the shuttle connection.\end{trrow}
\begin{trrow}{TrPlan}\trstep{TrPlan}{Comparison.} The reference answer is E. From the shared state, \care{} adds one observation and answers correctly, while GRPO adds none and answers incorrectly.\end{trrow}
\end{trajbox}
\caption{Full route-planning comparison underlying Figure~\ref{fig:case-main}(a). Red marks the reference region, cyan the shared observation and orange the additional \care{} observation.}
\label{fig:case-route-full}
\end{figure}

\begin{figure}[p]
\centering
\begin{trajbox}{ReseeSuccess}{Spatial Grounding: Commitment After Both Objects Are Located}
\includegraphics[width=\linewidth]{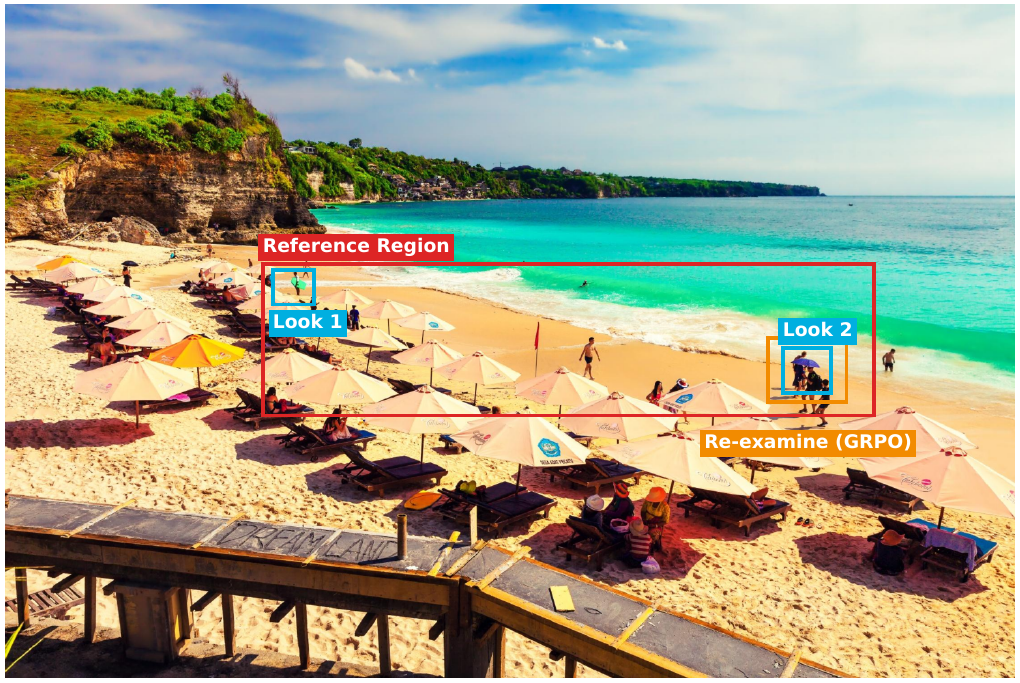}
\begin{trrow}{TrAsk}\trstep{TrAsk}{Question.} Is the green surfboard on the left or right side of the purple umbrella? (A) Right. (B) Left.\end{trrow}
\begin{trrow}{TrPlan}\trstep{TrPlan}{Initial Scene.} The beach contains many umbrellas, people and boards. The requested relation requires grounding both the green surfboard and the purple umbrella, rather than comparing arbitrary nearby objects.\end{trrow}
\begin{trrow}{TrLook}\trstep{TrLook}{Shared Observation 1.} Inspect the person carrying a green surfboard near the left side of the shoreline. The crop verifies the board's identity and color. Its position is retained in the original image.\end{trrow}
\begin{trrow}{TrLookB}\trstep{TrLookB}{Shared Observation 2.} Inspect the purple umbrella carried by the group farther right along the shoreline. The crop distinguishes this target from the many stationary pale umbrellas. Both object identities and their original horizontal order are now resolved.\end{trrow}
\begin{trrow}{TrOk}\trstep{TrOk}{CARE Verdict and Answer.} Commit to B, left. The green surfboard is to the left of the purple umbrella in the original scene. No additional observation is requested.\end{trrow}
\begin{trrow}{TrRex}\trstep{TrRex}{GRPO Verdict and Observation.} Re-examine the purple umbrella. The new crop returns to the same object and confirms the same color and location, adding no new relation needed for the answer.\end{trrow}
\begin{trrow}{TrOk}\trstep{TrOk}{GRPO Answer.} Commit to B, left, after the repeated inspection.\end{trrow}
\begin{trrow}{TrPlan}\trstep{TrPlan}{Comparison.} Both answers match the reference B. From the shared state, \care{} adds no observation and GRPO adds one. The difference concerns stopping after the evidence is complete.\end{trrow}
\end{trajbox}
\caption{Full spatial-grounding comparison underlying Figure~\ref{fig:case-main}(b). Red marks the reference region, cyan the shared observations and orange GRPO's repeated inspection.}
\label{fig:case-beach-full}
\end{figure}

Figures~\ref{fig:case-route-full} and~\ref{fig:case-beach-full} expand the two comparisons in Figure~\ref{fig:case-main}. Each includes the original scene, the evidence available at the shared decision state, and both policies' continuations through the final answer. Additional observation counts start at that shared state.

\subsection{Additional Trajectories}
\begin{figure}[p]
\centering
\begin{trajbox}{ReseeSuccess}{Success: counting bicycles under occlusion, nine turns}
\includegraphics[width=\linewidth]{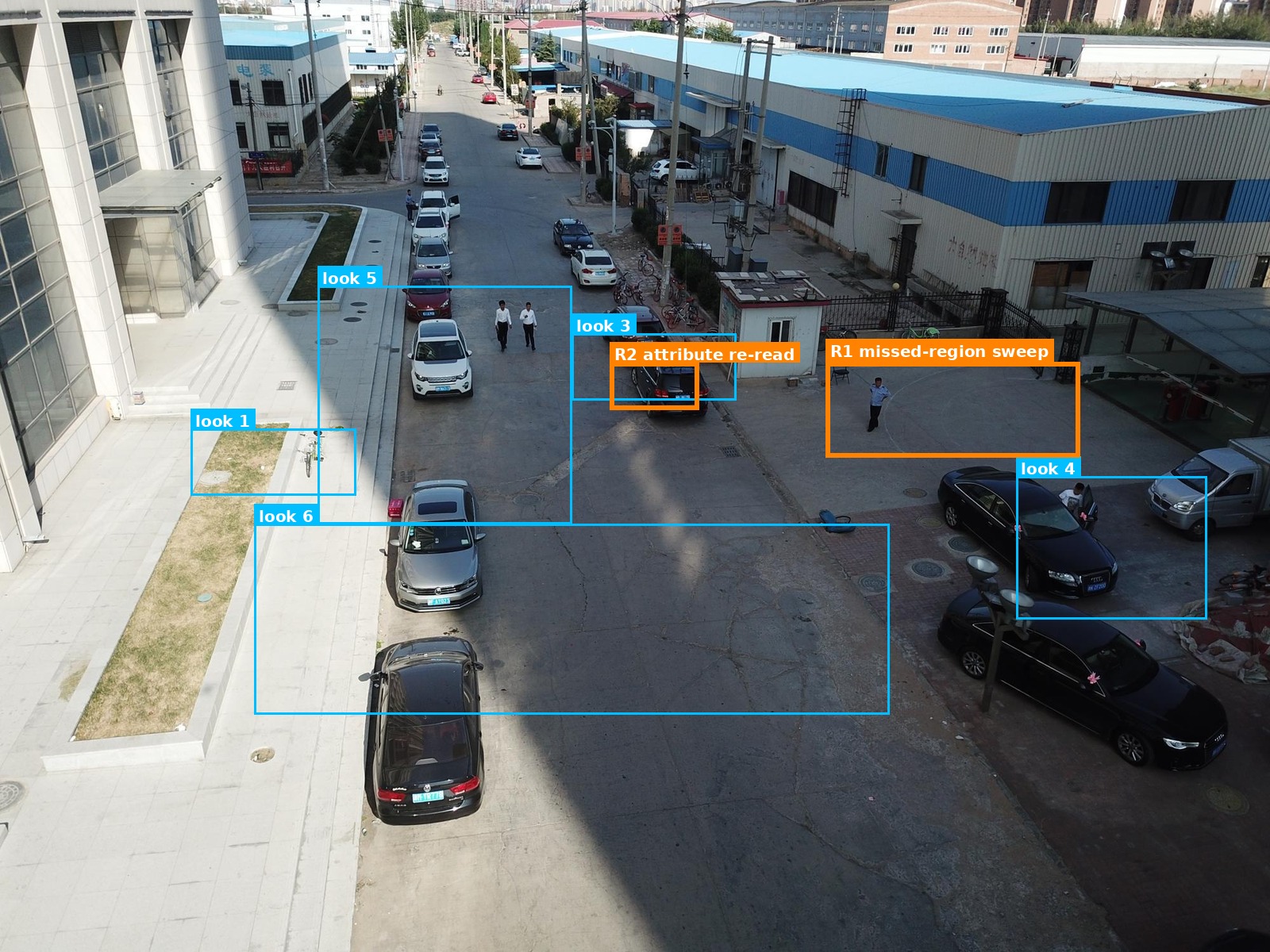}
\vspace{3pt}
\begin{trrow}{TrAsk}\trstep{TrAsk}{Question.} How many bicycles are in the image? \enspace (A) 9 \enspace (B) 22 \enspace (C) 18 \enspace (D) 16 \enspace (E) none.\end{trrow}
\begin{trrow}{TrPlan}\trstep{TrPlan}{Plan.} Bicycles are small next to the vehicles and buildings and may be occluded or clustered along the fences, so the left pavement, the right gate area and the cluster by the guardhouse are inspected in turn.\end{trrow}
\begin{trrow}{TrLook}\trstep{TrLook}{Looks 1--3.} A green bicycle stands beside the pavement. Behind the metal fence at the gate, four bicycles are parked in a row and two more lean against the posts, six in that zone. By the guardhouse another five are visible, including a red and a black one. The running tally is twelve.\end{trrow}
\begin{trrow}{TrLookB}\trstep{TrLookB}{Looks 4--5.} The right pavement and the left row of parked cars contain none.\end{trrow}
\begin{trrow}{TrRex}\trstep{TrRex}{Re-examine 1.} \textit{Missed-region sweep.} The gate zone was counted at six while the total is still short of every option, and the earlier crop stopped at the fence line. Re-scanning it finds a black-framed bicycle behind the silver car and another against the far wall, so that zone holds eight.\end{trrow}
\begin{trrow}{TrLookB}\trstep{TrLookB}{Look 6.} The guardhouse zone is confirmed at five.\end{trrow}
\begin{trrow}{TrRexB}\trstep{TrRexB}{Re-examine 2.} \textit{Attribute re-read.} Fourteen are counted and the nearest options are sixteen and eighteen, so the shadowed area is read again. A bicycle is partly hidden under the rear wheel of the brown SUV and the front wheel of another shows behind a white car, which brings the count to sixteen.\end{trrow}
\begin{trrow}{TrOk}\trstep{TrOk}{Answer: D, 16.} Both re-examinations are warranted. The tally before them would have answered fourteen, which is not among the options.\end{trrow}\vspace{2pt}
\end{trajbox}
\caption{A counting question answered after two re-examinations.}
\label{fig:traj-count}
\end{figure}

\begin{figure}[p]
\centering
\begin{trajbox}{ReseeSuccess}{Success: separating two motion attributes, three turns}
\includegraphics[width=\linewidth]{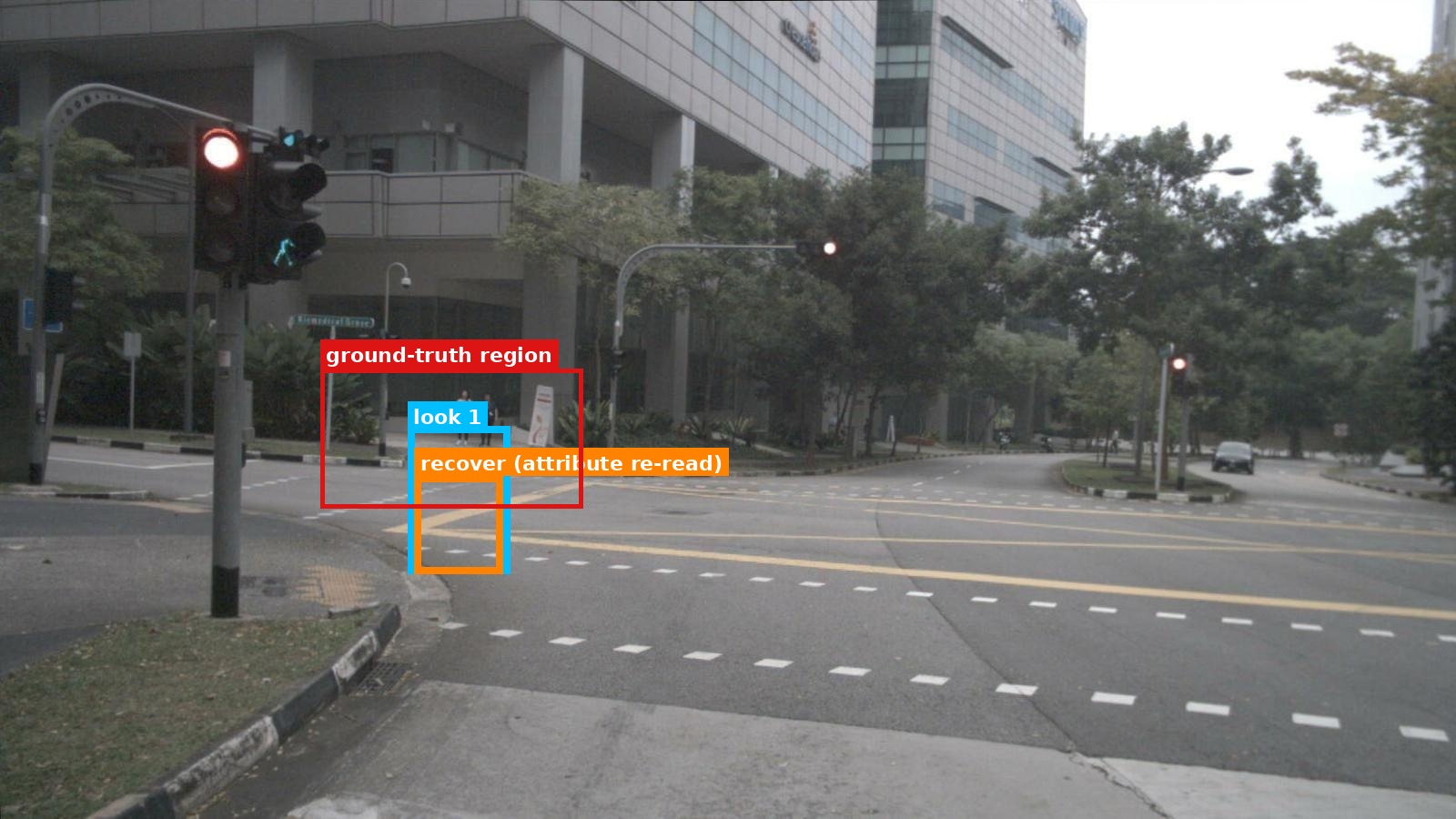}
\vspace{3pt}
\begin{trrow}{TrAsk}\trstep{TrAsk}{Question.} The image is the front view of the ego car. Predict its behavior. \enspace (A) going straight, fast \enspace (B) steering right, very fast \enspace (C) slightly steering right, very fast \enspace (D) slightly steering left, normal speed \enspace (E) none.\end{trrow}
\begin{trrow}{TrPlan}\trstep{TrPlan}{Plan.} The lane position, the curvature of the markings and the motion blur carry the answer. The lateral offset already points to a rightward steer, and the gentle curve suggests a slight one, which separates option C from option B.\end{trrow}
\begin{trrow}{TrLook}\trstep{TrLook}{Decision.} \textit{Verdict recover, risk medium.} One constraint is left open, whether the offset counts as a slight or a strong right steer. The two remaining options agree on the speed and differ only in that magnitude, so the region is read again before committing.\end{trrow}
\begin{trrow}{TrRex}\trstep{TrRex}{Re-examine 1.} \textit{Attribute re-read.} At higher resolution the left marking sits about half a meter from the bumper and the right one about four tenths, and the centerline is offset by roughly eight percent of the lane width. The asymmetry is small, which is mild corrective steering rather than an aggressive maneuver.\end{trrow}
\begin{trrow}{TrOk}\trstep{TrOk}{Answer: C.} One observation localizes the lane, and the re-examination measures the quantity that separates the two surviving options.\end{trrow}\vspace{2pt}
\end{trajbox}
\caption{A motion-attribute question settled by re-reading the lane markings.}
\label{fig:traj-driving}
\end{figure}

\begin{figure}[p]
\centering
\begin{trajbox}{ReseeFail}{Failure: the anchor entity is never grounded}
\includegraphics[width=\linewidth]{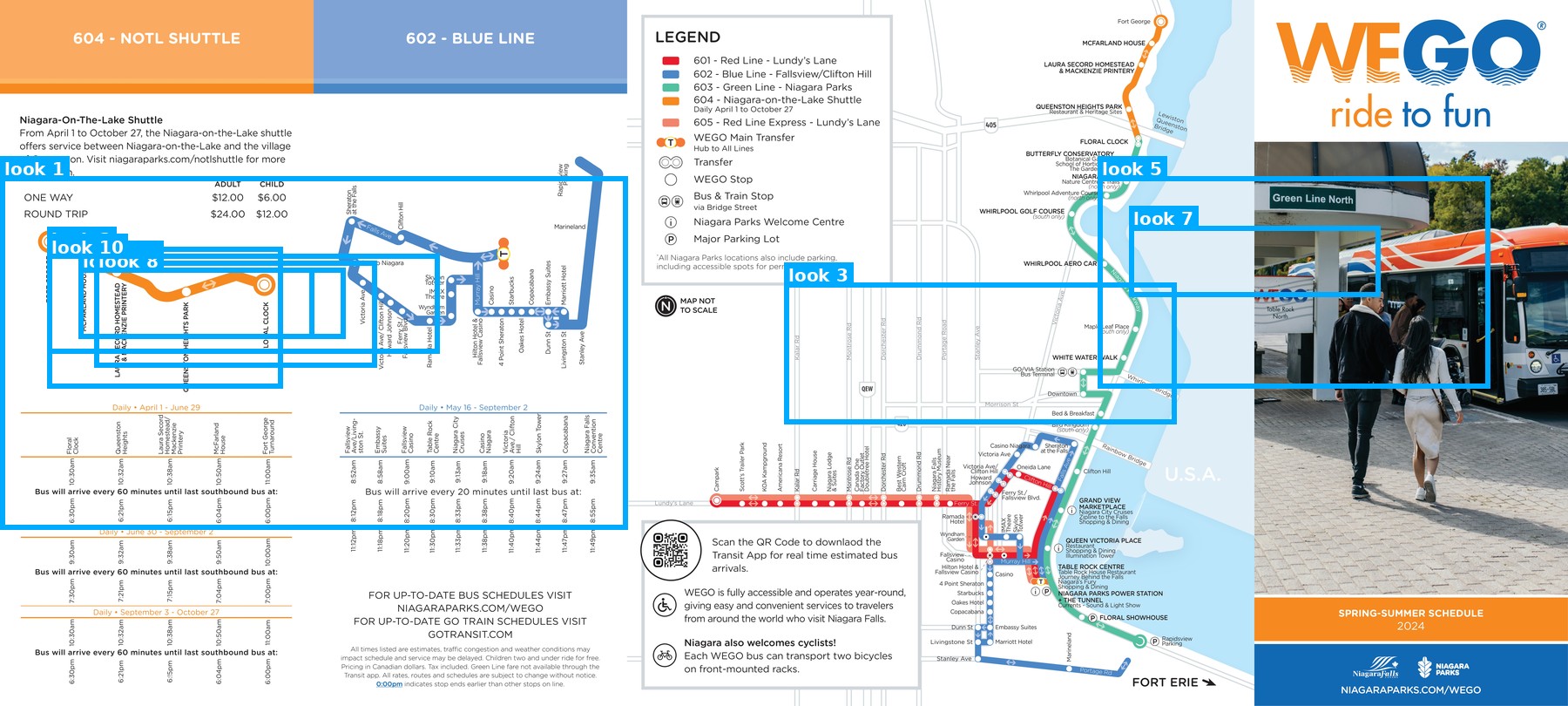}
\vspace{3pt}
\begin{trrow}{TrAsk}\trstep{TrAsk}{Question.} Which major parking lot is nearest to the Niagara Falls History Museum? The answer is a labeled parking icon beside a labeled terminus.\end{trrow}
\begin{trrow}{TrLook}\trstep{TrLook}{Looks 1--2.} The first crop covers the casino stops and reports that no museum label appears there, although a parking icon sits just west of one of them. The second finds a second parking icon by the bridge, and again no museum label.\end{trrow}
\begin{trrow}{TrLookB}\trstep{TrLookB}{Looks 3--10.} Seven further crops sweep the central corridor, the terminal and the eastern river edge. Every observation repeats the same finding, that the museum is not labeled in the region examined.\end{trrow}
\begin{trrow}{TrRex}\trstep{TrRex}{Decision.} \textit{Verdict recover, on each of these turns.} The gate reports that evidence is still missing, and every re-examination it triggers returns to the same corridor.\end{trrow}
\begin{trrow}{TrPlan}\trstep{TrPlan}{Commitment.} The model then reasons from outside the image, stating that the museum stands at Table Rock House and that its primary parking is Rapidsview, and treats that as confirmation.\end{trrow}
\begin{trrow}{TrBad}\trstep{TrBad}{Answer: C, incorrect.} The reference answer is A. Ten observations are spent confirming that a label is absent, and the question is finally resolved from prior knowledge of the city rather than from the map.\end{trrow}\vspace{2pt}
\end{trajbox}
\caption{A failure in which the re-examinations never leave the region that lacks the evidence.}
\label{fig:traj-grounding}
\end{figure}

\begin{figure}[p]
\centering
\begin{trajbox}{ReseeFail}{Failure: a confident commitment after the wrong region}
\includegraphics[width=\linewidth]{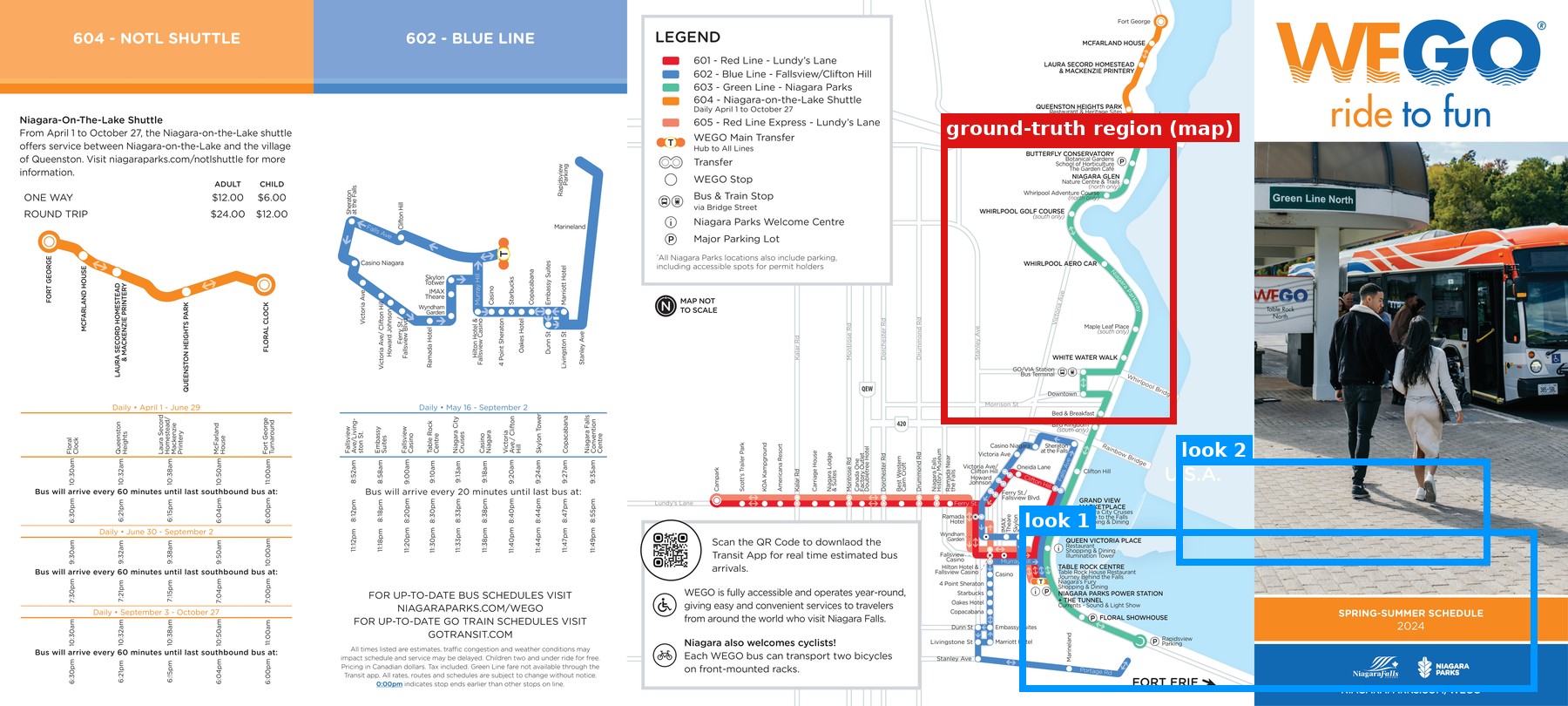}
\vspace{3pt}
\begin{trrow}{TrAsk}\trstep{TrAsk}{Question.} A traveler starts at the Marriott Hotel and ends at Queenston Heights Park without walking between transfers. How should they travel? The reference answer is E.\end{trrow}
\begin{trrow}{TrLook}\trstep{TrLook}{Look 1.} The crop reports the hotel labels and enumerates the stops of one line, then reasons about where two lines might meet.\end{trrow}
\begin{trrow}{TrLookB}\trstep{TrLookB}{Look 2.} A transfer hub is reported where the two lines intersect, with the destination park on the second of them.\end{trrow}
\begin{trrow}{TrRex}\trstep{TrRex}{Decision.} \textit{Verdict skip, risk low.} No constraint is left open. The transfer is described as visually confirmed, neither line is said to serve the hotel directly, and no other path is thought to exist.\end{trrow}
\begin{trrow}{TrBad}\trstep{TrBad}{Answer: D, incorrect.} Both crops land on the photographic inset in the lower right of the map rather than on the route diagram, so the confirmed transfer hub and the park label are read from a region that contains neither. The gate reports low risk precisely because the mislocalization was never noticed.\end{trrow}\vspace{2pt}
\end{trajbox}
\caption{A failure in which the model commits with low risk after reading the wrong region.}
\label{fig:traj-skip}
\end{figure}

Figures~\ref{fig:traj-count} to~\ref{fig:traj-skip} follow four trajectories of the trained policy on the evaluation split from the first observation to the answer, two successful and two unsuccessful. Each overlay marks the requested regions in order, in blue for an observation and orange for a re-examination.

\section{Additional Results}
\label{app:results}
Table~\ref{tab:coldstart} shows that pruning unwarranted demonstrations raises GRPO macro accuracy by $0.6$ points, while \care{} from the full cold start exceeds pruned GRPO by $1.9$ points. Masking unwarranted actions while retaining their surrounding trajectories preserves most of \care{}'s gain, consistent with learning from search context while correcting which re-examinations to imitate.

\begin{table}[t]
\centering
\caption{Cold-start variants evaluated before and after reinforcement learning.}
\label{tab:coldstart}
\resizebox{\textwidth}{!}{%
\begin{tabular}{l c ccc ccc ccc}
\toprule
 & & \multicolumn{3}{c}{\textbf{Cold start only}} & \multicolumn{3}{c}{\textbf{+ GRPO}} & \multicolumn{3}{c}{\textbf{+ \care{}}} \\
\cmidrule(lr){3-5}\cmidrule(lr){6-8}\cmidrule(lr){9-11}
\textbf{Variant} & Re-ex.\ kept & Acc.\ $\uparrow$ & Prec.\ $\uparrow$ & Over.\ $\downarrow$ & Acc.\ $\uparrow$ & Prec.\ $\uparrow$ & Over.\ $\downarrow$ & Acc.\ $\uparrow$ & Prec.\ $\uparrow$ & Over.\ $\downarrow$ \\
\midrule
\rowcolor{ReseeBand}\multicolumn{11}{l}{\textit{Label-free}} \\
Full & 450 &\textbf{39.3}&22.1&23.5&\underline{46.0}&27.3&13.6&\textbf{48.5}&\textbf{46.8}&10.6\\
Stripped & 0 &37.0&18.2&25.4&43.9&24.5&19.1&45.1&34.7&15.3\\
Single-turn & 450 &37.9&20.2&26.0&44.5&25.2&16.5&46.0&37.8&13.1\\
External & 0 &34.7&17.9&\underline{12.0}&42.8&23.7&\textbf{9.0}&44.0&32.0&\textbf{8.1}\\
\rowcolor{ReseeBand}\multicolumn{11}{l}{\textit{Label-assisted}} \\
Pruned & 95 &38.8&\textbf{35.0}&\textbf{10.2}&\textbf{46.6}&\textbf{37.1}&\underline{9.3}&47.4&\underline{45.1}&\underline{8.9}\\
Masked & 450 &\underline{39.1}&\underline{29.0}&16.7&45.9&\underline{31.8}&11.5&\underline{48.1}&45.0&9.8\\
\bottomrule
\end{tabular}}
\end{table}

\begin{table}[t]
\centering
\caption{Per-source answer accuracy for the ablations in Table~\ref{tab:ablation}. Macro is the unweighted mean of the six sources.}
\label{tab:ablation-sources}
\setlength{\tabcolsep}{4pt}
\small
\resizebox{\textwidth}{!}{%
\begin{tabular}{l cccccc c}
\toprule
\textbf{Variant} & V* & HR-Bench & MME-RW & TreeBench & O3-Bench & VisualProbe & \textbf{Macro} \\
\midrule
\care{} &\textbf{66.8}&\textbf{68.1}&\textbf{59.9}&\textbf{39.0}&\textbf{35.1}&\textbf{21.9}&\textbf{48.5}\\
\emph{w/o} localization &65.1&66.8&57.5&36.7&31.7&17.0&45.8\\
\emph{w/o} verdict flip &64.5&65.7&55.8&35.0&29.0&15.2&44.2\\
\emph{w/o} cost &65.6&66.5&57.0&36.0&31.0&16.3&45.4\\
w/ trajectory cost &66.1&67.3&58.3&37.3&32.6&17.4&46.5\\
\emph{w/o} verdict entropy &\underline{66.3}&\underline{67.5}&\underline{59.0}&\underline{38.1}&\underline{33.5}&\underline{18.8}&\underline{47.2}\\
\bottomrule
\end{tabular}}
\end{table}

\begin{table}[t]
\centering
\caption{Sensitivity of Qwen3-VL-8B to the four principal \care{} parameters. Each block varies one parameter with the others at their main-experiment values.}
\label{tab:sensitivity}
\small
\begin{tabular}{l c c c c}
\toprule
Parameter & Value & Macro acc. $\uparrow$ & Bal. acc. $\uparrow$ & Looks $\downarrow$ \\
\midrule
Credit weight $\beta$ & 0.25 &45.8&55.8&2.11\\
 & 0.5 &47.3&61.0&2.04\\
\rowcolor{ReseeBlue!10} & 1 &\textbf{48.5}&\textbf{65.9}&\textbf{1.97}\\
 & 2 &\underline{47.8}&\underline{62.4}&\underline{2.03}\\
 & 4 &46.0&56.8&2.13\\
\midrule
Clip $\kappa$ & 0.25 &46.7&59.6&2.04\\
 & 0.5 &\underline{47.9}&\underline{63.2}&\underline{1.99}\\
\rowcolor{ReseeBlue!10} & 1 &\textbf{48.5}&\textbf{65.9}&\textbf{1.97}\\
 & 2 &47.6&61.4&2.08\\
\midrule
Entropy $\eta$ & 0 &47.2&61.6&2.05\\
 & 0.003 &\underline{48.0}&\underline{64.0}&\underline{2.00}\\
\rowcolor{ReseeBlue!10} & 0.01 &\textbf{48.5}&\textbf{65.9}&\textbf{1.97}\\
 & 0.03 &47.9&63.5&2.04\\
 & 0.1 &46.1&56.7&2.16\\
\midrule
\rowcolor{ReseeBlue!10}Replays $k$ & 1 &48.5&65.9&1.97\\
 & 2 &\underline{48.8}&66.3&\underline{1.96}\\
 & 4 &\textbf{49.0}&\underline{66.6}&\textbf{1.95}\\
 & 8 &\textbf{49.0}&\textbf{66.7}&\textbf{1.95}\\
\bottomrule
\end{tabular}
\end{table}

\begin{table}[t]
\centering
\caption{Utility ranking on 288 unseen decisions, including 52 warranted re-examinations. Higher AUC indicates better discrimination between warranted and unwarranted decisions.}
\label{tab:utility}
\small
\setlength{\tabcolsep}{9pt}
\begin{tabular}{l c}
\toprule
\textbf{Decision signal} & \textbf{AUC} $\uparrow$ \\
\midrule
Self-assessed answer-flip risk &0.55\\
Unresolved-constraint count &0.54\\
\midrule
\care{}, $k=1$ &0.69\\
\care{}, $k=4$ &\underline{0.72}\\
\care{}, $k=8$ &\textbf{0.73}\\
\bottomrule
\end{tabular}
\end{table}

\begin{table}[t]
\centering
\caption{Observation-price sweeps for Qwen3-VL-8B. Every row is separately trained. \care{} prices the verdict contrast, while GRPO prices the trajectory reward.}
\label{tab:cost}
\small
\setlength{\tabcolsep}{8pt}
\begin{tabular}{l c c c c}
\toprule
Method & $\lambda$ & Macro acc. $\uparrow$ & Bal. acc. $\uparrow$ & Looks $\downarrow$ \\
\midrule
\multirow{4}{*}{\care{}} & $0$ &45.4&54.7&2.14\\
 & $2\times10^{-4}$ &\textbf{48.5}&\textbf{65.9}&1.97\\
 & $10^{-3}$ &\underline{47.8}&\underline{63.0}&1.86\\
 & $5\times10^{-3}$ &45.1&59.6&\textbf{1.70}\\
\midrule
\multirow{4}{*}{GRPO} & $0$ &46.0&54.8&2.09\\
 & $2\times10^{-4}$ &46.2&55.6&2.04\\
 & $10^{-3}$ &45.8&55.7&1.97\\
 & $5\times10^{-3}$ &44.4&54.5&\underline{1.73}\\
\bottomrule
\end{tabular}
\end{table}

\end{document}